%% file: main.tex
\documentclass[runningheads]{llncs}

\usepackage{eccv}

\usepackage{eccvabbrv}

\usepackage{graphicx}
\usepackage{booktabs}

\usepackage[accsupp]{axessibility}  % Improves PDF readability for those with disabilities.

\usepackage{hyperref}

\usepackage{orcidlink}

\usepackage{indentfirst}
\usepackage{tabularx}
\usepackage{multirow}
\usepackage{fontawesome}
\usepackage{marvosym}
\usepackage{colortbl}
\usepackage{soul}

\usepackage{marvosym}
\begin{document}

\title{MART: MultiscAle Relational Transformer Networks for Multi-agent Trajectory Prediction} 

\titlerunning{MART: Multiscale Relational Transformer Networks}

\author{Seongju Lee\inst{1}\orcidlink{0000-0002-1712-0282} \and
Junseok Lee\inst{1}\orcidlink{0000-0001-5212-2657} \and
Yeonguk Yu\inst{1}\orcidlink{0000-0003-2147-4718} \and
Taeri Kim\inst{1}\orcidlink{0000-0002-5139-7963} \and
\\Kyoobin Lee\inst{1}\thanks{Corresponding author.}\orcidlink{0000-0003-4299-4923}}

\authorrunning{Lee et al.}

\institute{School of Integrated Technology (SIT), Gwangju Institute of Science and Technology
(GIST), Cheomdan-gwagiro 123, Buk-gu, Gwangju 61005, Republic of Korea \email{\{lsj2121, junseoklee, yeon\_guk, smterry0928, kyoobinlee\}@gist.ac.kr}}

\maketitle

\begin{abstract}
    Multi-agent trajectory prediction is crucial to autonomous driving and understanding the surrounding environment. Learning-based approaches for multi-agent trajectory prediction, such as primarily relying on graph neural networks, graph transformers, and hypergraph neural networks, have demonstrated outstanding performance on real-world datasets in recent years. However, the hypergraph transformer-based method for trajectory prediction is yet to be explored. Therefore, we present a \textbf{M}ultisc\textbf{A}le \textbf{R}elational \textbf{T}ransformer (\textbf{MART}) network for multi-agent trajectory prediction. MART is a hypergraph transformer architecture to consider individual and group behaviors in transformer machinery. The core module of MART is the encoder, which comprises a Pair-wise Relational Transformer (PRT) and a Hyper Relational Transformer (HRT). The encoder extends the capabilities of a relational transformer by introducing HRT, which integrates hyperedge features into the transformer mechanism, promoting attention weights to focus on group-wise relations. In addition, we propose an Adaptive Group Estimator (AGE) designed to infer complex group relations in real-world environments. Extensive experiments on three real-world datasets (NBA, SDD, and ETH-UCY) demonstrate that our method achieves state-of-the-art performance, enhancing ADE/FDE by 3.9\%/11.8\% on the NBA dataset. Code is available at \url{https://github.com/gist-ailab/MART}.
\end{abstract}

\section{Introduction}
\label{sec:intro}
Multi-agent trajectory prediction, which forecasts the future trajectories of agents based on their past movements, holds significance in understanding human behavior and has applications in several fields, including robotics \cite{zhu2021learning, karnan2022socially} and autonomous driving \cite{chai2020multipath, li2019grip}. This task is complex and challenging for several reasons. For instance, some pedestrians move in formation, often as part of a group, introducing interdependencies and shared decision-making. These group dynamics influence individual paths, making predicting the trajectory of a single agent without considering the broader context challenging. Consequently, researchers have explored sophisticated methods to capture individual interactions and group behavior \cite{rudenko2018human, xu2022groupnet, bae2022gpgraph, zhou2012understanding, xu2022dynamic}.

\begin{figure}[t!]
  \includegraphics[width=\columnwidth]{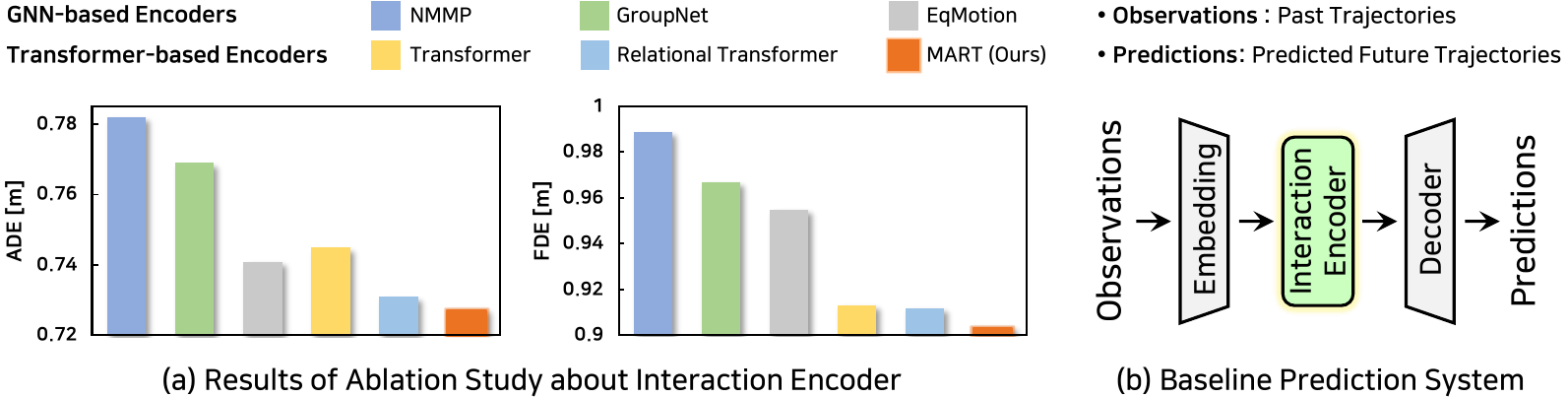}
  \caption{\textbf{(a) Results of ablation study about different interaction encoders on the NBA dataset. (b) Baseline prediction system used in ablation study.} We conduct an ablation study on interaction encoders to compare the MART encoder with state-of-the-art encoders. The average displacement error (ADE) and final displacement error (FDE) are assessed using the NBA dataset.}
  \label{fig:concept}
\end{figure}

Previous studies have focused on modeling pair-wise interactions in trajectory prediction using graph neural networks (GNNs) \cite{mohamed2020social, sun2020recursive, bae2021disentangled, shi2021sgcn, huang2019stgat, kosaraju2019social, liang2020garden, bae2022non, hu2020collaborative, xu2023eqmotion} and transformer networks \cite{vaswani2017attention, yu2020spatio, yuan2021agentformer, gu2022stochastic, mao2023leapfrog}. Some studies have employed graph convolutional networks \cite{kipf2017semisupervised} to account for pair-wise relations in crowded scenes \cite{mohamed2020social, sun2020recursive, bae2021disentangled, shi2021sgcn}. Other methods, such as those based on graph attention networks \cite{veličković2018graph}, have been proposed in \cite{huang2019stgat, kosaraju2019social, liang2020garden, bae2022non}. In addition, message-passing neural networks \cite{gilmer2017neural} have been employed in trajectory prediction tasks \cite{hu2020collaborative, xu2022groupnet, xu2022dynamic}. Recently, Xu \etal \cite{xu2023eqmotion} introduced a motion prediction model under Euclidean geometric transformations based on equivariant GNNs \cite{satorras2021n}. In addition, Yu \etal \cite{yu2020spatio} proposed a graph transformer network that considers pair-wise and temporal relations by incorporating an adjacency matrix into the attention mechanism.

To cooperate with individual and group behaviors, GroupNet \cite{xu2022groupnet}, a multiscale hypergraph message-passing neural network, was proposed for multi-agent trajectory prediction. As an extension of GroupNet \cite{xu2022groupnet}, DynGroupNet \cite{xu2022dynamic} captures time-varying interactions at the pair and group scales. However, transformer machinery, an effective relationship-learning mechanism, is not fully incorporated into their methods. Bae \etal \cite{bae2022gpgraph} proposed the GP-Graph, which considers intra- and inter-group relationships by partitioning the most likely behavioral agents in the same group. However, round-trip node-edge communication, crucial for relational reasoning \cite{diao2023relational}, is ignored in GP-Graph.

In this study, we propose a \textbf{M}ultisc\textbf{A}le \textbf{R}elational \textbf{T}ransformer (\textbf{MART}) network, a hypergraph transformer architecture for multi-agent trajectory prediction. To address social relationships in hypergraph transformer architecture, we introduce the MART encoder (MARTE), comprising a pair-wise relational transformer and a Hyper Relational Transformer (HRT) to enhance the capabilities of the relational transformer \cite{diao2023relational}. We introduce Hyper Relational Attention (HRA) into HRT, incorporating hyperedge features into the attention mechanism. As hyperedge features involve group information, HRT captures group behavior while promoting attention weights to focus on group-wise relations.

Since group relations are not explicitly defined in real-world scenarios, pioneering studies have introduced methods for reasoning group relations \cite{bae2022gpgraph, xu2022groupnet}. However, they have limitations. In \cite{bae2022gpgraph}, the focus is on non-overlapping group relations that cannot handle complex relationships. For \cite{xu2022groupnet}, the number of agents in a group requires a manual definition, and as the number of group scales increases, more encoders are required, resulting in higher computational costs. To address the aforementioned limitations, we propose an Adaptive Group Estimator (AGE). Inspired by the Straight-Through Estimation (STE) technique \cite{bengio2013estimating} used in binarized neural networks \cite{hubara2016binarized, liu2018bi, xu2019accurate, wang2020sparsity}, we utilize adaptive thresholding to include highly correlated agents within the same group. Our approach estimates overlapping group relations, recognizing that agents can be correlated with more than one group. Moreover, unlike \cite{xu2022groupnet}, the manual definition of the number of agents in a group is unnecessary in our method.

To validate the effectiveness of MART, we conduct extensive experiments on three real-world datasets (NBA, SDD \cite{robicquet2016learning}, and ETH-UCY \cite{pellegrini2009you, lerner2007crowds}) and demonstrate that our method achieves state-of-the-art (SOTA) performance. Notably, on the NBA dataset, which is the most challenging and complex, MART shows a subjective performance improvement by reducing the ADE by 3.9\% and FDE by 11.8\% compared to the previous SOTA method, EqMotion \cite{xu2023eqmotion}. The ablation study on the interaction encoder demonstrates that hypergraph transformer architecture is beneficial for multi-agent trajectory prediction. The proposed MARTE, based on a hypergraph transformer, outperforms other SOTA GNN-based and transformer-based encoders, as illustrated in Figure \ref{fig:concept}. Additionally, MART exhibits computational efficiency compared to recent SOTA models. The main contributions are as follows:

\begin{itemize}
    \item We introduce MART, a Multiscale RelAtional Transformer network for multi-agent trajectory prediction. MART captures relations at individual and group scales through a hypergraph transformer mechanism via MARTE. The proposed MARTE promotes the attention weights to emphasize interactions between agents, particularly focusing on group-wise relations.
    \item We propose the AGE module to infer group relations. The AGE module outperforms existing group reasoning methods in both performance and efficiency by considering overlapping group relations and dealing with complex interactions through adaptive thresholding.
    \item We experimentally demonstrate that MART achieves SOTA performance on three real-world datasets (NBA, SDD, and ETH-UCY). Moreover, MART is computationally efficient compared to the recent SOTA models.
\end{itemize}

\section{Related Work}

\noindent\textbf{Trajectory Prediction.}\hspace{0.4em}
Traditional trajectory prediction methods are based on heuristics, energy potentials, and simple machine learning-based methods~\cite{mehran2009abnormal, berndt2008continuous, morris2011trajectory, hu2004learning}. Berndt~\etal\cite{berndt2008continuous} proposed using the global vehicle position with a hidden Markov model to predict object trajectory in a road traffic environment. Subsequently, recurrent neural network (RNN)-based methods \cite{zyner2018recurrent, altche2017lstm, vemula2018social} have been proposed to capture the long-term behavior of agents. For instance, Vemula~\etal. \cite{vemula2018social} used an RNN-based prediction model with attention that captured the relative importance of each person to predict agents' trajectories in the crowd. Generative adversarial network (GAN)-based methods \cite{dendorfer2021mg, fang2020tpnet, gupta2018social, hu2020collaborative, sadeghian2019sophie, sun2020reciprocal} and variational autoencoder-based methods \cite{lee2022muse, mangalam2020not, salzmann2020trajectron++, xu2022groupnet, xu2022dynamic} have also been studied to generate future trajectory distribution. Similarly, \cite{gu2022stochastic, mao2023leapfrog} proposed trajectory-prediction methods based on a diffusion model. In addition, transformer structures \cite{giuliari2021transformer, yu2020spatio, yuan2021agentformer} have been used to capture the importance of the agents' spatial and temporal correlations. Our method is a hypergraph transformer network that captures individual and group behaviors for trajectory prediction.

\noindent\textbf{Transformers with Graph.}\hspace{0.4em}
Recently, transformer-based structures have been proposed for representing graphs \cite{diao2023relational, min2022transformer, chen2022structure, yu2020spatio}. Specifically, Chen~\etal\cite{chen2022structure} addressed concerns regarding transformer architecture, in which the node representation generated by the transformer fails to capture the structural similarity between them, and solved it by incorporating structural information into the self-attention module. In addition, relational transformer \cite{diao2023relational}, which uses an attention module to capture edge information effectively, has been demonstrated to perform diverse graph-structure tasks. For trajectory prediction, Li~\etal\cite{li2022graph} proposed a graph-based spatial transformer to capture the spatial interactions between pedestrians. Herein, we introduce MART to effectively capture individual and group behaviors via transformer mechanisms.

\section{Background and Problem Formulation}
\subsection{Trajectory Prediction}
Multi-agent trajectory prediction predicts the future trajectories of multiple agents based on their past trajectories. Assume that $N$ agents exist in a multi-agent system. Specifically, let $\mathbf{X} \in \mathbb{R}^{N \times T_p \times 2}$, $\mathbf{Y} \in \mathbb{R}^{N \times T_f \times 2}$, and $\Hat{\mathbf{Y}} \in \mathbb{R}^{N \times T_f \times 2}$ represent past trajectories, ground-truth future trajectories, and predicted future trajectories in 2D spatial coordinates, respectively. The objective is to minimize the error between the predicted and ground-truth future trajectories. We design a trajectory prediction model denoted by $\Hat{\mathbf{Y}} = \mathcal{M}(\mathbf{X})$. Here, $\mathcal{M}$ produces predicted future trajectories that closely align with the ground-truth future trajectories.

\subsection{Relational Transformer}\label{subsec:rt}
The relational transformer (RT) \cite{diao2023relational}, an extension of the transformer \cite{vaswani2017attention}, incorporates edge vectors as fundamental components to achieve SOTA performance in various tasks involving graph-structured data. Mathematically, for a given input-directed graph, we denote an attributed graph as $\mathcal{G}=(\mathcal{N}, \mathcal{E})$. Here, $\mathcal{N}$ represents an unordered set of node vectors $\mathbf{n}_i\in\mathbb{R}^{d_n}$, where $\mathbf{n}_i$ is the $i$-th node vector and $d_n$ is the node dimension. $\mathcal{E}$ represents the set of edge vectors $\mathbf{e}_{ij}\in\mathbb{R}^{d_e}$, where $\mathbf{e}_{ij}$ represents the directed edge pointing from $j$-th to $i$-th node and $d_e$ is the edge dimension.

Similar to message-passing neural networks \cite{gilmer2017neural}, RT comprises two processes: node and edge updates. In the node update, RT introduces relational attention (RA), which incorporates edge vectors into the scaled dot-product attention. Formally, the query, key, and value vectors in RA are defined as follows:
\begin{equation}
\begin{array}{ll}
    \mathbf{q}_{ij} = (\mathbf{n}_{i}\mathbf{W}_{n}^{Q} + \mathbf{e}_{ij}\mathbf{W}_{e}^{Q}) \in \mathbb{R}^{d_n},\\
    \mathbf{k}_{ij} = (\mathbf{n}_{j}\mathbf{W}_{n}^{K} + \mathbf{e}_{ij}\mathbf{W}_{e}^{K}) \in \mathbb{R}^{d_n},\\
    \mathbf{v}_{ij} = (\mathbf{n}_{j}\mathbf{W}_{n}^{V} + \mathbf{e}_{ij}\mathbf{W}_{e}^{V}) \in \mathbb{R}^{d_n},
\end{array}
\end{equation}
where $\mathbf{W}_n\in\mathbb{R}^{d_n \times d_n}$ and $\mathbf{W}_e\in\mathbb{R}^{d_e \times d_n}$ are the weight matrices used to project the node and edge vectors onto the query, key, and value vectors, respectively. The remainder of the node update is the same as the transformer mechanism. Formally, the updated node vector after $(l+1)$-th layer $\mathbf{n}_{i}^{(l+1)}\in\mathbb{R}^{d_n}$ is defined as follows:
\begin{equation}
    \mathbf{n}_i^{(l+1)} = \mathcal{F}_n\left(\mathrm{softmax}_{j} \left( \frac{\mathbf{q}_{ij}^{(l)}{\mathbf{k}^{(l)}_{ij}}^{\top}}{\sqrt{d_n}} \right) \mathbf{v}_{ij}^{(l)}, \mathbf{n}_i^{(l)}\right)
\end{equation}
where $\mathcal{F}_n(\cdot)$ represents a transformer-like node update function of the form [Add \& Norm]–[FeedForward]–[Add \& Norm]. In this study, we omitted the equation for multi-head attention for simplicity.

For the edge update, the edge vector after $(l+1)$-th layer $\mathbf{e}_{ij}^{(l+1)}\in\mathbb{R}^{d_e}$ is produced by the edge update function $\mathcal{F}_e(\cdot)$ as follows:
\begin{equation}\label{eq:ra_edge_update}
    \mathbf{e}_{ij}^{(l+1)}=\mathcal{F}_e\left(\mathbf{m}_{ij}^{(l)}, \mathbf{e}_{ij}^{(l)}\right).
\end{equation}
Here, $\mathbf{m}^{(l)}_{ij} \in\mathbb{R}^{d_e}$ represents the message vector that aggregates the information from adjacent updated nodes and their edges. Function $\mathcal{F}_e(\cdot)$ denotes the transformer-like edge-update function, which has an identical operation as $\mathcal{F}_n(\cdot)$. The message vector $\mathbf{m}^{(l)}_{ij}$ is derived as follows:
\begin{equation}
\mathbf{m}^{(l)}_{ij} = \mathrm{ReLU}([\mathbf{e}_{ij}^{(l)};\mathbf{e}_{ji}^{(l)};\mathbf{n}_{i}^{(l+1)};\mathbf{n}_{j}^{(l+1)}]\mathbf{W}_m),
\end{equation}
where $[\cdot ; \cdot]$ denotes the concatenation operation and $\mathbf{W}_m \in \mathbb{R}^{(2d_e + 2d_n)\times d_{h}}$ is a weight matrix for the message function.

\begin{figure*}[t]
\begin{center}
  \includegraphics[width=\columnwidth]{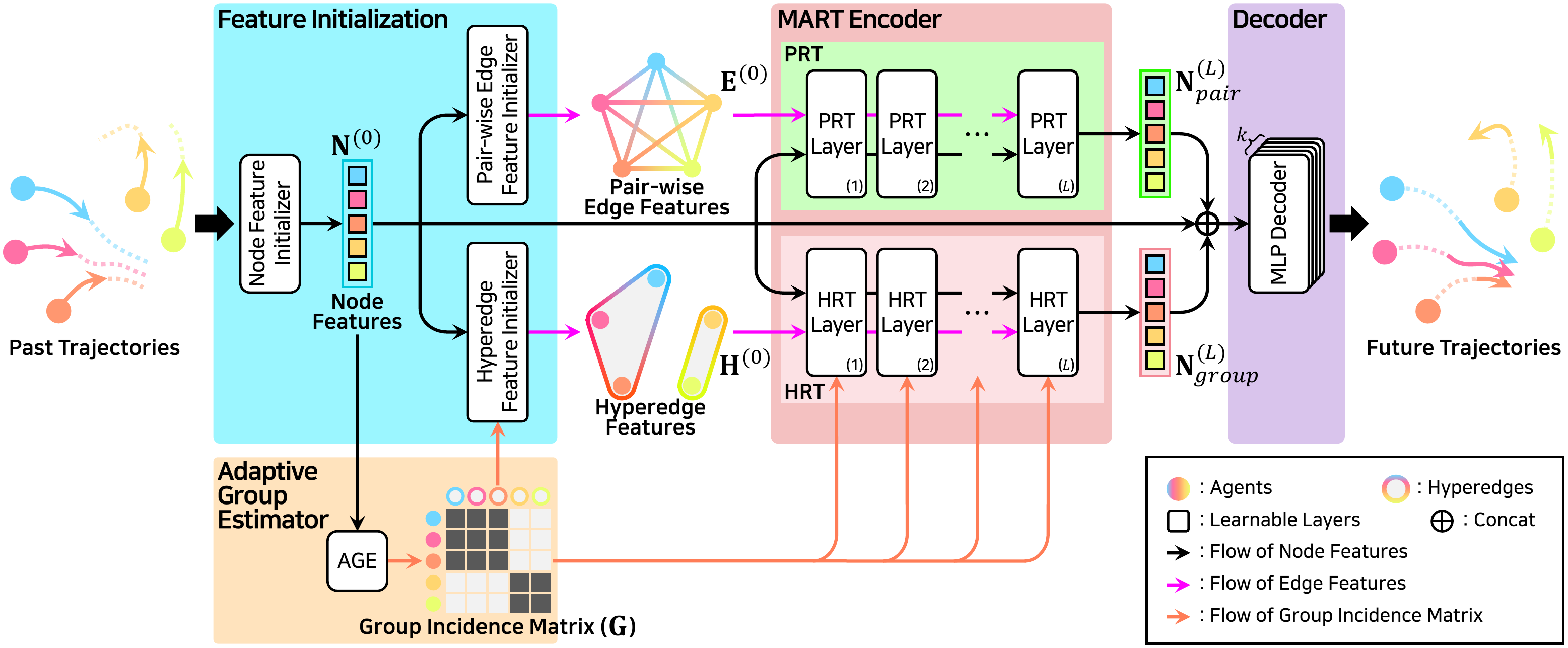}
  \caption{\textbf{Model architecture of the proposed MultiscAle Relational Transformer (MART) network.} Our model has four components: feature initialization (\textcolor[HTML]{00E2FA}{sky blue}), Adaptive Group Estimator (\textcolor[HTML]{FFA731}{orange}), MART encoder (\textcolor[HTML]{EE6969}{red}), and future trajectory decoder (Decoder) (\textcolor[HTML]{B287D9}{purple}). The MART encoder includes a pair-wise relational transformer (\textcolor[HTML]{29F600}{green}) and a hyper relational transformer (\textcolor[HTML]{F38794}{pink}).}
  \label{fig:main_model}
\end{center}
\end{figure*}

\section{Method}
As illustrated in Figure \ref{fig:main_model}, MART comprises four components: feature initialization, AGE, MARTE, and decoder. The subsequent sections offer a detailed description of the core components of MART—feature initialization, AGE module, and MARTE—as well as the overall process.

\subsection{Feature Initialization} \label{subsec:feature_initialization}
We first transform the past input trajectories into agent embeddings to represent the initial node and edge features. Because edge relationships are often not explicitly defined in real-world environments, we represent edge features by aggregating the features of the adjacent nodes. In this paper, we consider pair-wise edge and hyperedge features.

\vspace{0.5em}
\noindent\textbf{Node Feature Initialization.}\hspace{0.4em}
The past trajectory of $i$-th agent $\mathbf{X}_i\in\mathbb{R}^{T_p \times d_i}$ is embedded via $\mathbf{n}^{(0)}_{i} = \mathcal{F}_{\mathrm{NI}}(\mathbf{X}_{i})$, where $\mathbf{n}_{i}^{(0)}\in\mathbb{R}^{d_n}$ is the initial node feature of the $i$-th agent and $\mathcal{F}_{\mathrm{NI}}(\cdot)$ represents two layers of multilayer perceptron (MLP). $d_n$ denotes the node dimensions.

\vspace{0.5em}
\noindent\textbf{Edge Feature Initialization.}\hspace{0.4em}
To consider pair- and group-wise relations, we initialize pair-wise edge and hyperedge features, respectively. To form pair-wise edge features, two adjacent nodes are concatenated and passed through the pair-wise edge feature initializer $\mathcal{F}_{\mathrm{PI}}(\cdot)$: $\mathbf{e}^{(0)}_{ij} = \mathcal{F}_{\mathrm{PI}}([\mathbf{n}^{(0)}_{i}; \mathbf{n}^{(0)}_{j}])$, where $\mathbf{e}^{(0)}_{ij}$ is the initial pair-wise edge feature from $j$-th to $i$-th agent and $\mathcal{F}_{\mathrm{PI}}(\cdot)$ is MLP.

The initial hyperedge feature is formed by averaging the initial node features of the corresponding hyperedge. The aggregated node features are fed to the hyperedge initializer $\mathcal{F}_{\mathrm{HI}}(\cdot)$: $\mathbf{h}^{(0)}_{i} = \mathcal{F}_{\mathrm{HI}}\left(\frac{1}{|\mathcal{N}_i|} \sum_{n_j\in \mathcal{N}_i} \mathbf{n}^{(0)}_{j} \right)$, where $\mathcal{F}_{\mathrm{HI}}$ is implemented by MLP, $\mathcal{N}_i$ is a set of node features of the $i$-th hyperedge, and $|\mathcal{N}_i|$ is its cardinality. In the following section, we propose a group reasoning method to address the lack of predefined hypergraph topology in real-world environments.

\subsection{Adaptive Group Estimator}
Real-world environments often conceal group relationships, which are crucial for understanding group behavior. Thus, several works have proposed methods for reasoning about group relations \cite{bae2022gpgraph, xu2022groupnet}; however, these methods have limitations, as mentioned in the Introduction. To overcome the aforementioned limitations, we propose the Adaptive Group Estimator (AGE) to effectively infer complex group relationships through adaptive thresholding, thereby grouping closely related agents.

Mathematically, given a hypergraph $\mathcal{G} = (\mathcal{N}, \mathcal{H})$, a group relation can be expressed as a group incident matrix $\mathbf{G}\in\mathbb{R}^{|\mathcal{N}| \times |\mathcal{H}|}$, where $\mathbf{G}_{ij}\in \{0, 1\}$. Here, $\mathcal{H}$ is an unordered set of hyperedge features, and $|\mathcal{H}|$ is its cardinality. When $\mathbf{G}_{ij}=1$, the $i$-th agent belongs to the $j$-th hyperedge; however, when $\mathbf{G}_{ij}=0$, it does not. In this study, we assume that the number of hyperedges is set to $|\mathcal{N}|$ to include agents highly correlated with the $j$-th agent (ego agent) in the same group. Thus, the $(i,j)$-th element of the group incidence matrix $\mathbf{G}$ is defined as follows:
\begin{equation}\label{eq:incidence_matrix}
    \mathbf{G}_{ij} = \mathcal{U}(\mathbf{A}_{ij} - \mathrm{\Theta}),
\end{equation}
where $\mathbf{A}_{ij} = \mathbf{n}_{i}^{\top}\mathbf{n}_{j}/(||\mathbf{n}_{i}||_{2}||\mathbf{n}_{j}||_{2})\in\mathbb{R}^{N \times N}$ is the $(i,j)$-th element of the affinity matrix, $\mathrm{\Theta} \in (-1.0, 1.0)$ is a learnable threshold, and $\mathcal{U}(\cdot)$ is a unit step function defined as follows:
\begin{equation}
    \mathcal{U}(x) = \begin{cases}
    1, & \text{if $x \geq 0$}.\\
    0, & \text{otherwise}.
  \end{cases}
\end{equation}

As the unit step function is not differentiable, we use the STE trick \cite{bengio2013estimating} inspired by binarized neural networks \cite{hubara2016binarized, liu2018bi, xu2019accurate, wang2020sparsity}. We compute the group incidence matrix $\mathbf{G}$ using Eq. \ref{eq:incidence_matrix} for the forward path. In the backward path, we estimate the gradient of $\mathcal{U}(\cdot)$ to enable backpropagation. The estimated gradient of $\mathcal{U}(\cdot)$ is defined as Eq. \ref{eq:partial_u}. Note that this estimation is derived based on Eq. \ref{eq:original_u}, which approximates $\mathcal{U}(\cdot)$. By introducing $\mathrm{\Theta}$ into the AGE module, it selectively groups more correlated agents into the same group. The supplementary material describes the details of the function selection.

\vspace{-2.0em}

\begin{figure*}
  \begin{minipage}[t]{.45\linewidth}
\vspace{1.35em}
\hspace{-0.5em}\begin{equation} \label{eq:partial_u}
\frac{\partial \mathcal{U}(x)}{\partial x} \approx \begin{cases}
2-4|x|, & \hspace{-0.5em}\text{if $|x| \leq 0.5$},\\
0, & \hspace{-0.5em}\text{otherwise}.
\end{cases}\hspace{-0.8em}
\end{equation}
  \end{minipage}\hfil
  \begin{minipage}[t]{.55\linewidth}
\begin{equation} \label{eq:original_u}
\mathcal{U}(x) \approx \begin{cases}
0, & \hspace{-0.5em}\text{if $x < -0.5$},\\
0.5+2x+2x^2, & \hspace{-0.5em}\text{if $-0.5 \leq x < 0$},\\
0.5+2x-2x^2, & \hspace{-0.5em}\text{if $0 \leq x < 0.5$},\\
1, & \hspace{-0.5em}\text{otherwise}.
\end{cases}\hspace{-0.8em}
\end{equation}
  \end{minipage}
\end{figure*}

\vspace{-2.0em}

\begin{figure*}
  \begin{minipage}[t]{.48\linewidth}
      \includegraphics[width=\linewidth]{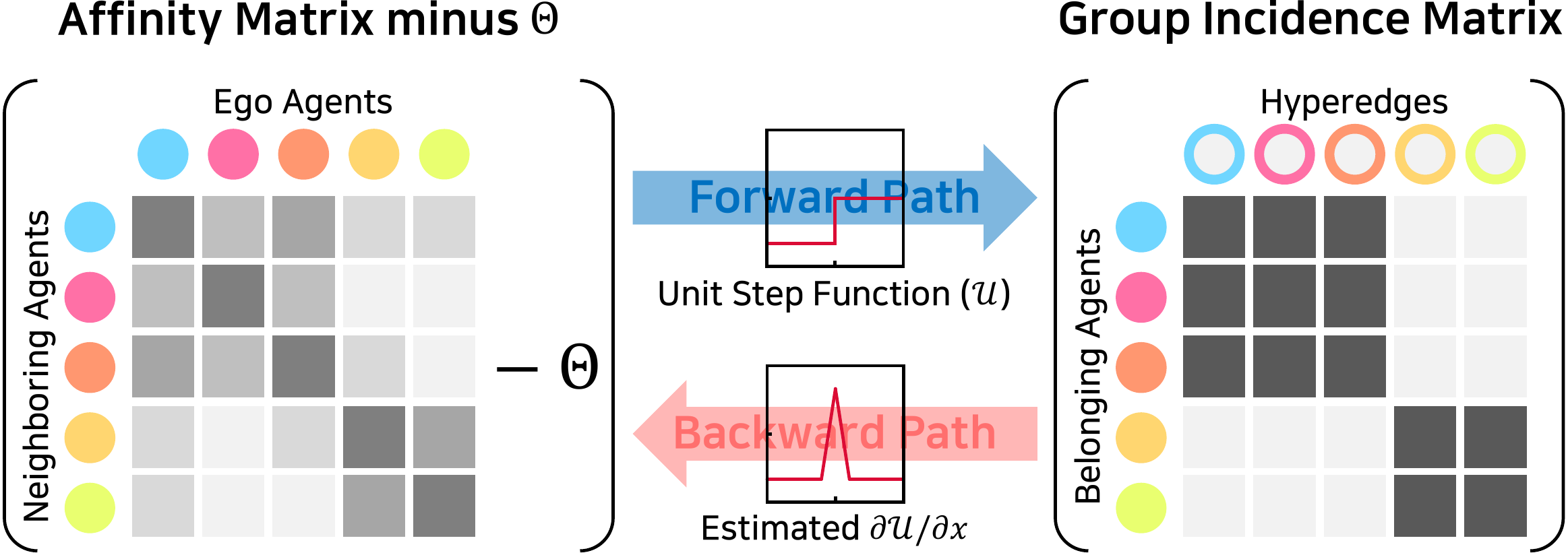}
      \caption{\textbf{Illustration of adaptive group estimator (AGE) module.} The AGE module employs an adaptive thresholding inspired by STE trick \cite{bengio2013estimating}. The \textcolor[HTML]{7FB7DF}{blue} and \textcolor[HTML]{FFB7B6}{red} arrows represent the forward and estimated backward paths, respectively.} 
      \label{fig:age_module}
  \end{minipage}\hfil
  \begin{minipage}[t]{.46\linewidth}
      \includegraphics[width=\linewidth]{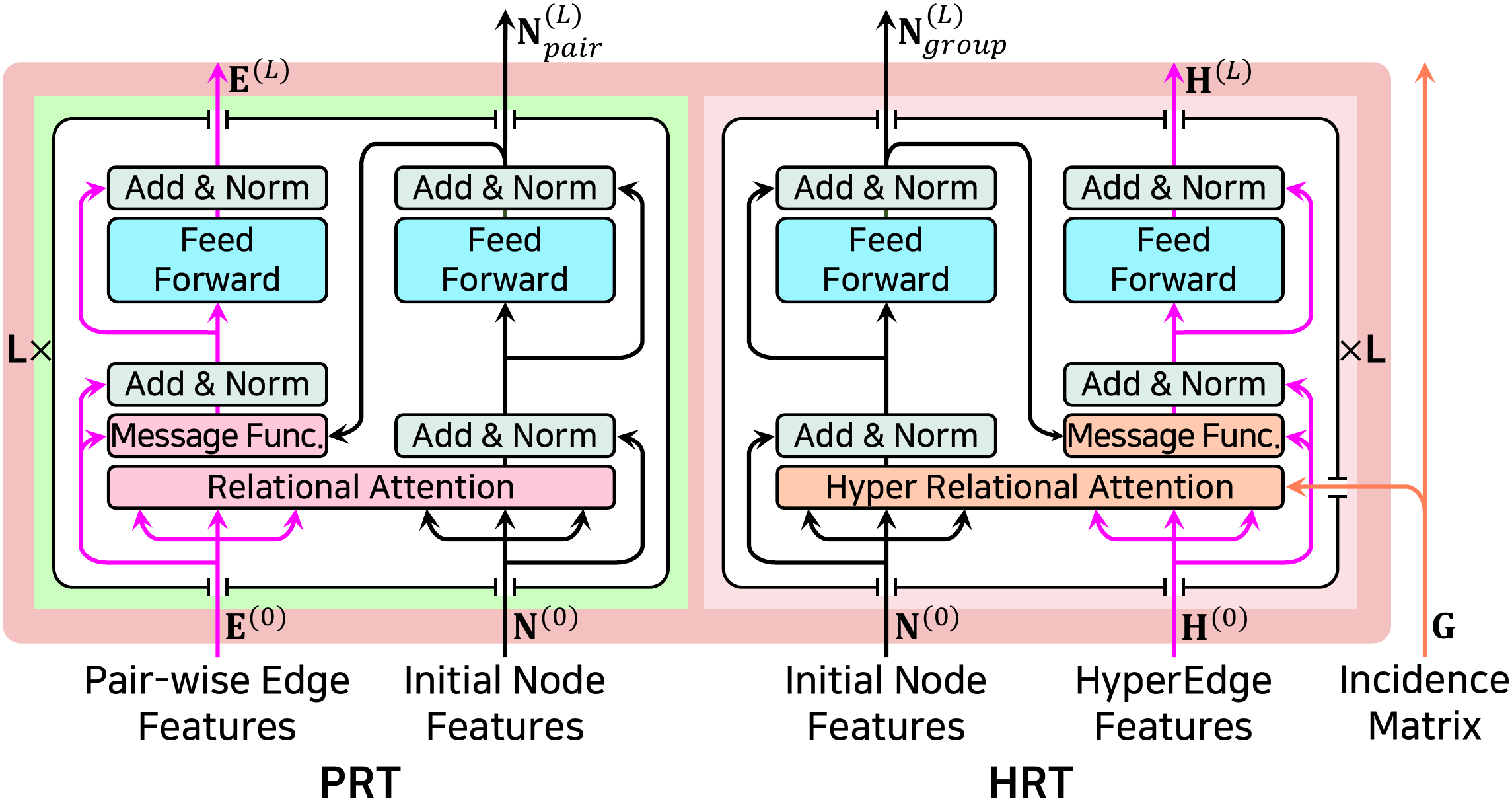}
      \caption{\textbf{Architecture of MARTE.} The regions in \textcolor[HTML]{29F600}{green}, \textcolor[HTML]{F38794}{pink}, and \textcolor[HTML]{EE6969}{red} correspond to those in Figure \ref{fig:main_model}. The black/\textcolor[HTML]{FF00FF}{purple}/\textcolor[HTML]{FF7F50}{coral} arrows denote the flow of node features/edge features/group incidence matrix.} 
      \label{fig:MARTE}
  \end{minipage}
\end{figure*}

\subsection{Multiscale Relational Transformer Encoder}\label{subsec:grt}
We introduce the MultiscAle Relational Transformer Encoder (MARTE) into our network to capture social interactions at multiple scales via hypergraph transformer architecture. The detailed architecture of MARTE is shown in Figure \ref{fig:MARTE}. PRT and HRT address individual and group behaviors, respectively.

\vspace{1.0em}
\noindent\textbf{Pair-wise Relational Transformer.}\hspace{0.4em}
PRT is based on RT \cite{diao2023relational}, an effective relational learner, as described in Section \ref{subsec:rt}. However, RT deals only with pair-wise relationships. Therefore, in the following section, we introduce HRT as an extension of RT to accommodate hypergraph structures, ensuring synergy between architectures.

\vspace{1.0em}
\noindent\textbf{Hyper Relational Transformer.}\hspace{0.4em}
We propose HRT to capture group behavior via a hypergraph transformer mechanism. To achieve this, we introduce Hyper Relational Attention (HRA) into HRT, extending the relational attention mechanism to accommodate hypergraph structures. Additionally, we introduce a message function to update the hyperedges in accordance with these structures.

(\textit{Node update for HRT}) To extend the mechanism of relational attention to a hypergraph structure, we integrate hyperedge information into the query, key, and value by aggregating the hyperedge features to which the corresponding agent belongs. The average operation is used for aggregation for stable training. \textit{The aggregated hyperedge features in query, key, and value work as the relative positional information that includes group information,} formulated as follows:

\begin{equation}
    \begin{array}{lll}
    \mathbf{q}_{i} = \left( \mathbf{n}_{i}\mathbf{W}_{n}^{Q} + {\displaystyle \frac{1}{|\mathcal{H}_i|}\sum_{h_{k}\in\mathcal{H}_{i}}}\mathbf{h}_{k}\mathbf{W}_{h}^{Q} \right) \in \mathbb{R}^{d_n},\\
    \mathbf{k}_{j} = \left( \mathbf{n}_{j}\mathbf{W}_{n}^{K} + {\displaystyle \frac{1}{|\mathcal{H}_j|}\sum_{h_{k}\in\mathcal{H}_{j}}}\mathbf{h}_{k}\mathbf{W}_{h}^{K} \right) \in \mathbb{R}^{d_n},\\
    \mathbf{v}_{j} = \left( \mathbf{n}_{j}\mathbf{W}_{n}^{V} + {\displaystyle \frac{1}{|\mathcal{H}_j|}\sum_{h_{k}\in\mathcal{H}_{j}}}\mathbf{h}_{k}\mathbf{W}_{h}^{V} \right) \in \mathbb{R}^{d_n},
    \end{array}
\end{equation}

where $\mathbf{W}_h\in\mathbb{R}^{d_e \times d_n}$ is the weight matrix used to project the hyperedge features onto the query, key, and value vectors, $\mathcal{H}_i$ is a set of hyperedges to which the $i$-th agent belongs, and $|\mathcal{H}_i|$ is its cardinality. These relations of belonging are derived from $\mathbf{G}$. Consequently, the updated node feature $\mathbf{n}_{i}^{(l+1)}$ is formulated as follows:
\begin{equation}
    \mathbf{n}_i^{(l+1)} = \mathcal{F}_n\left(\mathrm{softmax}_{j} \left( \frac{\mathbf{q}_{i}^{(l)}{\mathbf{k}^{(l)}_{j}}^{\top}}{\sqrt{d_n}} \right) \mathbf{v}_{j}^{(l)}, \mathbf{n}_i^{(l)}\right).
\end{equation}

(\textit{Hyperedge update for HRT}) The $i$-th hyperedge feature after the $(l+1)$-th HRT layer, denoted as $\mathbf{h}_{i}^{(l+1)} \in \mathbb{R}^{d_e}$, is defined as follows:
\begin{equation}
    \mathbf{h}_{i}^{(l+1)} = \mathcal{F}_{h}(\mathbf{m}_{i}^{(l)}, \mathbf{h}_{i}^{(l)})
\end{equation}
where $\mathcal{F}_h(\cdot)$ denotes the hyperedge update function based on feedforward network, and $\mathbf{m}_{i}^{(l)}$ is a message vector. For the message vector $\mathbf{m}_{i}^{(l)}$, we formulate a message function to accommodate the hyperedge structure. More specifically, this function is formulated using the hyperedge feature and the mean of the updated node features belonging to the corresponding hyperedge. Mathematically, the message vector $\mathbf{m}_{i}^{(l)} \in \mathbb{R}^{d_e}$ is defined as follows:
\begin{equation}
\mathbf{m}^{(l)}_{i} = \mathrm{ReLU}\left(\left[\mathbf{h}_{i}^{(l)};\frac{1}{|\mathcal{N}_i|}\sum_{n_{j} \in \mathcal{N}_i}\mathbf{n}_{j}^{(l+1)}\right]\mathbf{W}_m\right),
\end{equation}
where $\mathbf{W}_m\in\mathbb{R}^{(d_e + d_n) \times d_h}$ denotes the weight matrix to project post node-update features onto the message vector.

\subsection{Overall Process of MART}
We detail the trajectory prediction process of MART in the following sequence: feature initialization, encoding, decoding, and loss functions. The first three processes represent the forward path of MART \textit{in matrix formula}, and the loss function serves as the objective function for model training.

\vspace{1.0em}
\noindent\textbf{Feature Initialization.}\hspace{0.4em}
As described in Section \ref{subsec:feature_initialization}, our model first initializes the node, pair-wise edge, and hyperedge features for $N$ agents as:
\begin{equation}
\begin{array}{lll}
    \mathbf{N}^{(0)} = \{ \mathbf{n}^{(0)}_1, \mathbf{n}^{(0)}_2, ..., \mathbf{n}^{(0)}_N \}\in\mathbb{R}^{N \times d_n},\\
    \mathbf{E}^{(0)}=\{ \mathbf{e}^{(0)}_{11}, \mathbf{e}^{(0)}_{12}, ..., \mathbf{e}^{(0)}_{NN} \}\in\mathbb{R}^{N^2 \times d_e},\\
    \mathbf{H}^{(0)}=\{ \mathbf{h}^{(0)}_1, \mathbf{h}^{(0)}_2, ..., \mathbf{h}^{(0)}_N \}\in\mathbb{R}^{N \times d_e}.
\end{array}
\end{equation}
The pair-wise edge features $\mathbf{E}$ have $N^2 \times d_e$ dimensions because we consider a fully connected directed graph with self-loops.

\vspace{0.5em}
\noindent\textbf{Encoding Process.}\hspace{0.4em}
MARTE encodes pair- and group-wise node features from the initial features. Formally, the features produced by MARTE are described as follows:
\begin{equation}
\begin{array}{ll}
    \mathbf{N}^{(L)}_{pair} = \mathcal{PRT}(\mathbf{N}^{(0)}, \mathbf{E}^{(0)}),\\
    \mathbf{N}^{(L)}_{group} = \mathcal{HRT}(\mathbf{N}^{(0)}, \mathbf{H}^{(0)}, \mathbf{G}),
\end{array}\label{eq:L-th_node_feat}
\end{equation}
where $\mathbf{N}^{(L)}_{pair} \in \mathbb{R}^{N \times d_n}$ represents the node features at the pair-wise scale from the $L$-th PRT layer and $\mathbf{N}^{(L)}_{group}\in\mathbb{R}^{N \times d_n}$ denotes the node features at the group scale from the $L$-th HRT layer. We use $\mathcal{PRT}(\cdot)$ to refer to the PRT operation and $\mathcal{HRT}(\cdot)$ for the HRT operation.

\vspace{0.5em}
\noindent\textbf{Decoding Process.}\hspace{0.4em}
To predict future trajectories, we utilize multiple prediction heads as the decoder to obtain $k$ predictions, as in \cite{xu2023eqmotion}. Three types of features are concatenated: $\mathbf{N}^{(0)}$, $\mathbf{N}^{(L)}_{pair}$, and $\mathbf{N}^{(L)}_{group}$, then fed into the multi-head decoder $\mathcal{F}_{\mathrm{D}}$. Formally, the trajectories obtained from the $k$-th prediction head can be described as $\hat{\mathbf{Y}}_k = \mathcal{F}_{\mathrm{D}, k}([\mathbf{N}^{(0)};\mathbf{N}^{(L)}_{pair};\mathbf{N}^{(L)}_{group}]),$ where $\mathcal{F}_{\mathrm{D}, k}(\cdot)$ denotes a decoder with three layers of MLP.

\vspace{0.5em}
\noindent\textbf{Loss Function.}\hspace{0.4em}
We trained our model using the variety loss proposed in social-GAN \cite{gupta2018social} to optimize the best prediction. With the given $k$-th predicted trajectories resulting from $k$-th prediction head $\hat{\mathbf{Y}}_k$, the minimum of the L2-norm among $k$ predictions is utilized for backpropagation. Mathematically, the loss function is formulated as $\mathcal{L}=\frac{1}{NT_f} \sum_{n=1}^N\sum_{t=1}^{T_f} \min _{k}\left\|Y^{(n,t)} - \widehat{Y}_{k}^{(n,t)}\right\|_{2}$, where $(n,t)$ denotes the indices for the time step $t$ for the $n$-th agent.

\section{Experiments}

\noindent\textbf{Datasets.}\hspace{0.4em}
For the evaluation of MART, we utilize three datasets for multi-agent trajectory prediction: NBA SportVU (NBA), Stanford Drone (SDD) \cite{robicquet2016learning}, and ETH-UCY \cite{pellegrini2009you, lerner2007crowds}. Further details about the datasets and evaluation scheme are described in the supplementary material.

\noindent\textbf{Implementation Details.}\hspace{0.4em}
The node and edge dimensions, denoted by $d_n$ and $d_e$, respectively, are set to 64. Eight heads are used for multi-head attention. The number of layers in MARTE, denoted by $L$, is 4. For all datasets, our network is trained using the Adam optimizer \cite{kingma2014adam}. A learning rate decay of 0.5 is applied every 100 epochs. The network is trained using an NVIDIA GeForce RTX 3090. All our implementations are on PyTorch 2.1.0 \cite{paszke2019pytorch}. Further details about the implementation can be found in the supplementary material.

\noindent\textbf{Evaluation Metrics.}\hspace{0.4em}\label{subsec:evaluation_metrics}
To assess the trajectory prediction performance, we employed two metrics, $\min\mathrm{ADE}_{k}$ and $\min\mathrm{FDE}_{k}$, as in \cite{alahi2016social, gupta2018social, lee2017desire}. $\mathrm{ADE}$ and $\mathrm{FDE}$ represent the Euclidean distances between the predicted and actual trajectories, considering the entire path and endpoints, respectively. Consequently, $\min \mathrm{ADE}_{k}$ and $\min \mathrm{FDE}_{k}$ are the smallest values among $k$ computed $\mathrm{ADE}$s and $\mathrm{FDE}$s from the $k$ predicted trajectories.

\section{Results and Discussion}

\subsection{Comparison with State-Of-The-Art (SOTA) Methods}\label{subsec:comparison_sota}

\begin{table*}[t]
\setlength\tabcolsep{2.6pt} % let LaTeX compute intercolumn whitespace
\centering
\caption{Performance comparison on (a) NBA, (b) SDD, and (c) ETH-UCY datasets. The values denote $\min\mathrm{ADE}_{20}$/$\min\mathrm{FDE}_{20}$ in meters. \textbf{Bold}/\underline{underline} indicates the best/second-best result. * denotes that values are reproduced using the official implementation} \label{tab:nba_result}
\resizebox{\columnwidth}{!}{
\begin{tabular*}{1.4\columnwidth}{l|cccccccc|c}
\toprule
\multicolumn{10}{c}{\textbf{(a) NBA Dataset}} \\
\midrule
\multirow{2}{*}{Time} & STAR & GroupNet & MemoNet & MID & NPSN & DynGroupNet & LED & EqMotion* & \multirow{2}{*}{Ours} \\
 & \cite{yu2020spatio} & \cite{xu2022groupnet} & \cite{xu2022remember} & \cite{gu2022stochastic} & \cite{bae2022non} & \cite{xu2022dynamic} & \cite{mao2023leapfrog} & \cite{xu2023eqmotion} & \\
\midrule
4.0s & 1.13/2.01 & 0.96/1.30 & 1.25/1.47 & 0.96/1.27 & 1.31/1.79 & 0.89/1.13 & 0.81/1.10 & \underline{0.76}/\underline{1.02} & \textbf{0.73}/\textbf{0.90} \\
\bottomrule
\toprule
\multicolumn{10}{c}{\textbf{(b) SDD Dataset}} \\
\midrule
\multirow{2}{*}{Time} & PECNet & GroupNet & MemoNet & MID & NPSN & DynGroupNet & ET & LED & \multirow{2}{*}{Ours} \\
  & \cite{mangalam2020not} & \cite{xu2022groupnet} & \cite{xu2022remember} & \cite{gu2022stochastic} & \cite{bae2022non} & \cite{xu2022dynamic} & \cite{bae2023eigentrajectory} & \cite{mao2023leapfrog} &\\
\midrule
4.8s &  9.96/15.88 & 9.31/16.11 & 8.56/12.66 & 9.73/15.32 & 8.56/11.85 & 8.42/13.58 & \underline{8.05}/13.25 & 8.48/\textbf{11.66} & \textbf{7.43}/\underline{11.82}  \\
\bottomrule
\toprule
\multicolumn{10}{c}{\textbf{(c) ETH-UCY Dataset}} \\
\midrule
\multirow{2}{*}{Subset} & GroupNet & MemoNet & MID & NPSN & GP-Graph & EqMotion & ET & LED & \multirow{2}{*}{Ours} \\
 & \cite{xu2022groupnet} & \cite{xu2022remember} & \cite{gu2022stochastic} & \cite{bae2022non} & \cite{bae2022gpgraph} & \cite{xu2023eqmotion} & \cite{bae2023eigentrajectory} & \cite{mao2023leapfrog} & \\
\midrule
ETH & 0.46/0.73 & 0.40/0.61 & 0.39/0.66 & \underline{0.36}/0.59 & 0.43/0.63 & 0.40/0.61 & \underline{0.36}/\underline{0.53} & 0.39/0.58 & \textbf{0.35}/\textbf{0.47} \\
HOTEL & 0.15/0.25 & \textbf{0.11}/\textbf{0.17} & 0.13/0.22 & 0.16/0.25 & 0.18/0.30 & \underline{0.12}/\underline{0.18} & \underline{0.12}/0.19 &  \textbf{0.11}/\textbf{0.17} & 0.14/0.22 \\
UNIV & 0.26/0.49 & 0.24/0.43 & \textbf{0.22}/0.45 & \underline{0.23}/\textbf{0.39} & 0.24/\underline{0.42} & \underline{0.23}/0.43 & 0.24/0.43 & 0.26/0.43  & 0.25/0.45 \\
ZARA1 & 0.21/0.39 & \underline{0.18}/0.32 & \textbf{0.17}/0.30 & \underline{0.18}/0.32 & \textbf{0.17}/0.31 & \underline{0.18}/0.32 & 0.19/0.33 & \underline{0.18}/\textbf{0.26} & \textbf{0.17}/\underline{0.29} \\
ZARA2 & 0.17/0.33 & \underline{0.14}/0.24 & \textbf{0.13}/0.27 & \underline{0.14}/0.25 & 0.15/0.29 & \textbf{0.13}/\underline{0.23} & \underline{0.14}/0.24 & \textbf{0.13}/\textbf{0.22} & \textbf{0.13}/\textbf{0.22} \\
\midrule
AVG & 0.25/0.44 & \textbf{0.21}/0.35 & \textbf{0.21}/0.38 & \textbf{0.21}/0.36 & \underline{0.23}/0.39 & \textbf{0.21}/0.35 & \textbf{0.21}/\underline{0.34} & \textbf{0.21}/\textbf{0.33} & \textbf{0.21}/\textbf{0.33} \\
\bottomrule

\end{tabular*}
}

\end{table*}

The performance of the proposed model is compared with the SOTA methods on the NBA, SDD, and ETH-UCY datasets using two metrics: $\min\mathrm{ADE}_{20}$ and $\min\mathrm{FDE}_{20}$, which are described in Section \ref{subsec:evaluation_metrics}.

\noindent\textbf{NBA Dataset.}\hspace{0.4em} Table \ref{tab:nba_result}-(a) presents the performance comparison to eight SOTA methods on the NBA dataset. Compared to the previous best method EqMotion \cite{xu2023eqmotion}, our model reduces $\min\mathrm{ADE}_{20}$ and $\min\mathrm{FDE}_{20}$ at 4.0 s by 3.9\% and 11.8\%, respectively.

\noindent\textbf{SDD Dataset.}\hspace{0.4em} We compared our method with eight SOTA methods on the SDD dataset, as listed in Table \ref{tab:nba_result}-(b). Our method enhances the $\min\mathrm{ADE}_{20}$ by 7.8\% compared to EigenTrajectory (ET) \cite{bae2023eigentrajectory} and shows comparable performance in $\min\mathrm{FDE}_{20}$ compared to LED \cite{mao2023leapfrog}.

\noindent\textbf{ETH-UCY Dataset.}\hspace{0.4em} Performance comparison with eight SOTA methods on the ETH-UCY dataset is listed in Table \ref{tab:nba_result}-(c). Our model yields results comparable to those of a previous SOTA method, LED \cite{mao2023leapfrog}.

\noindent\textbf{Computational Efficiency of MART.} \hspace{0.4em} The computational efficiency of MART is evaluated in comparison with recent SOTA methods in terms of model parameters and multiple-add cumulation (MAC) for inference. These metrics are computed with 10 agents in the ETH-UCY setting. As shown in Table \ref{tab:result_computation}, our method demonstrates computational efficiency compared to previous SOTA methods such as LED \cite{mao2023leapfrog} and EqMotion \cite{xu2023eqmotion}. Despite STAR \cite{yu2020spatio} having only 1.0M model parameters, it has a larger MAC count than ours due to its iterative architecture.

\noindent\textbf{Discussion.}\hspace{0.4em}
MART demonstrates SOTA or comparable performance in trajectory prediction compared to previous methods. Our approach is more computationally efficient, with 7.3x fewer parameters and 347.3x fewer MACs than the previous SOTA method, LED \cite{mao2023leapfrog}. These results indicate that MART is highly competitive for accurate trajectory prediction in real-time applications.

\subsection{Ablation Study}

\begin{table}[t]
    \begin{minipage}{.49\linewidth}
      \caption{Comparison of the computational efficiency of the proposed and state-of-the-art methods. \textbf{Bold}/\underline{underline} indicate the best/second-best. All values are obtained using the official implementation. * denotes an improved version from their GitHub. R2O refers to Ratio-to-Ours.} \label{tab:result_computation}
      \setlength\tabcolsep{5.0pt}
      \centering
        \resizebox{\columnwidth}{!}{
        \begin{tabular*}{1.4\columnwidth}{c|cc|cc}
        \toprule
        \textbf{Method} & \#Param.$\downarrow$ & R2O & \#MAC$\downarrow$ & R2O \\
        \midrule
        PECNet \cite{mangalam2020not} & 2.1M & 1.4x & 259.2M & 6.0x \\
        STAR \cite{yu2020spatio} & \textbf{1.0M} & \textbf{0.7x} & 12.0G & 276.9x \\
        MemoNet \cite{xu2022remember} & 10.7M & 7.1x & 6.0G & 137.8x \\
        GroupNet \cite{xu2022groupnet} & 2.2M & 1.5x & 411.5M & 9.5x \\
        MID* \cite{gu2022stochastic} & 9.0M & 6.0x & 40.3G & 931.7x \\
        EqMotion \cite{xu2023eqmotion} & 3.0M & 2.0x & \underline{147.1M} & \underline{3.4x} \\
        LED \cite{mao2023leapfrog} & 10.9M & 7.3x & 15.0G & 347.3x \\
        \midrule
        \textbf{MART (Ours)} & \underline{1.5M} & \underline{1.0x} & \textbf{43.3M} & \textbf{1.0x} \\
        \bottomrule
        \end{tabular*}
        }

    \end{minipage}%
    \hspace{0.4em}
    \begin{minipage}{.49\linewidth}
    \caption{Ablation studies of interaction encoders and group estimation modules on the NBA dataset. \textbf{Bold}/\underline{underline} indicate the best/second-best.} \label{tab:ablation}
    \setlength\tabcolsep{5.5pt}
    \centering
        \resizebox{\columnwidth}{!}{
        \begin{tabular*}{1.4\columnwidth}{c|cc}
        \toprule
        \textbf{Interaction Encoder} & $\min\mathrm{ADE}_{20}$ & $\min\mathrm{FDE}_{20}$ \\
        \midrule
        NMMP \cite{hu2020collaborative} & 0.782 & 0.989 \\
        GroupNet \cite{xu2022groupnet} & 0.769 & 0.967 \\
        EqMotion \cite{xu2023eqmotion} & 0.741 & 0.955 \\
        Transformer \cite{vaswani2017attention} & 0.745 & 0.913 \\
        Relational Transformer \cite{diao2023relational} & \underline{0.731} & \underline{0.912} \\
        \midrule
        \textbf{MARTE (ours)} & \textbf{0.727} & \textbf{0.903} \\
        \bottomrule
        \multicolumn{2}{c}{\vspace{-0.7mm}} \\
        \toprule
        \textbf{Group Estimation} & $\min\mathrm{ADE}_{20}$ & $\min\mathrm{FDE}_{20}$ \\
        \midrule
        Module in GP-Graph \cite{bae2022gpgraph} & 0.740 & 0.913 \\
        Module in GroupNet \cite{xu2022groupnet} & \underline{0.732} & \underline{0.912} \\
        \midrule
        \textbf{AGE (ours)} & \textbf{0.727} & \textbf{0.903} \\
        \bottomrule
        \end{tabular*}
        }
    \end{minipage} 
\end{table}

\noindent\textbf{Effect of MARTE.}\hspace{0.4em}
Ablation studies are conducted on different interaction encoders to demonstrate the effectiveness of MARTE, which is a core component of MART. Five interaction encoders are evaluated in our prediction system for a fair comparison. The model specifications, such as the number of layers and heads for transformer \cite{vaswani2017attention} and relational transformer \cite{diao2023relational}, are the same as those for MARTE. As illustrated at the top of Table \ref{tab:ablation}, MARTE outperforms both transformer-based encoders (transformer and relational transformer) and GNN-based encoders (NMMP, GroupNet, and EqMotion). On the other hand, the results demonstrate that transformer-based encoders perform better than GNN-based encoders. Although the relational transformer exhibits commendable performance, it overlooks the importance of group-wise interaction, which is essential for accurate trajectory prediction. Thus, the findings underscore that MARTE enhances performance via a hypergraph transformer, marking a novel approach in trajectory prediction.

\noindent\textbf{Effect of AGE.}\hspace{0.4em}
To validate our AGE module, it is compared to previously proposed group estimation methods \cite{xu2022groupnet, bae2022gpgraph} by integrating them into MART. As shown at the bottom of Table \ref{tab:ablation}, the AGE module performs better than the other methods. The group estimation of GP-Graph \cite{bae2022gpgraph} leads to lower performance in complex scenes like the NBA datasets due to non-overlapping group estimation. GroupNet \cite{xu2022groupnet} performs similarly to our method in ADE but requires manual group size setting. Moreover, as the number of group scales increases, more encoders are required, resulting in higher computational costs.

\subsection{Qualitative Results} \label{subsec:qualitative_results}

\begin{figure*}[t]
  \includegraphics[width=\columnwidth]{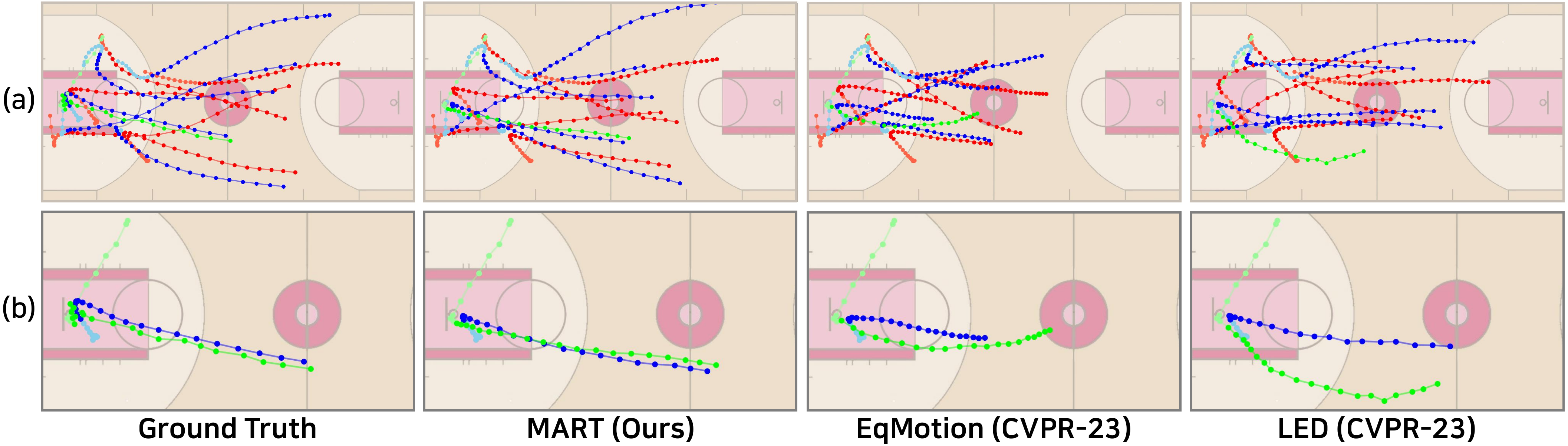}
  \caption{\textbf{Qualitative result comparison on NBA dataset.} We qualitatively compare our method with two recent SOTA methods using the best of 20 predictions. (a) illustrates predictions for all agents, and (b) focuses on the ball-keeping player and basketball. Our method predicts more accurate future trajectories compared to LED \cite{mao2023leapfrog} and EqMotion \cite{xu2023eqmotion}. Notably, as illustrated in (b), MART provides robust predictions about the ball-keeping player and basketball, while other methods do not. These results demonstrate that MART can handle more complex interactions compared to state-of-the-art (SOTA) methods. (Light color represents the past trajectory, while \textcolor[HTML]{0000FF}{blue}/\textcolor[HTML]{FF0000}{red}/\textcolor[HTML]{00FF00}{green} indicate the two teams and basketball.)}
  \label{fig:visualization}
\end{figure*}

\noindent\textbf{Trajectory Prediction.}\hspace{0.4em}
In Figure \ref{fig:visualization}, the predicted trajectories from MART (ours) are compared with two recent SOTA methods, LED \cite{mao2023leapfrog} and EqMotion \cite{xu2023eqmotion}, along the ground truth (GT) trajectories on the NBA dataset. MART outperforms the previous methods, resulting in accurate trajectory predictions. Further visualization can be found in the supplementary material.

\begin{figure}[t]
  \includegraphics[width=\columnwidth]{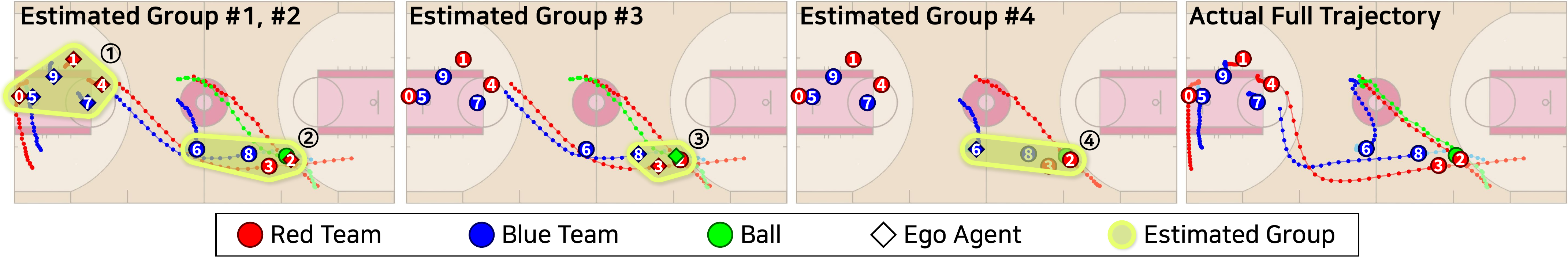}
  \caption{\textbf{Examples of group estimation results with predictions (first three figures) and actual full trajectories (last figure) on the NBA dataset.} Duplicated groups are excluded, and the light color denotes the past trajectory. The left side of the court represents the \textcolor{blue}{blue} team's region, while the right side of the court represents the \textcolor{red}{red} team's region. Group IDs are denoted as \textcircled{\raisebox{-0.9pt}{1}}, \textcircled{\raisebox{-0.9pt}{2}}, \textcircled{\raisebox{-0.9pt}{3}}, and \textcircled{\raisebox{-0.9pt}{4}}. Player IDs are provided in the figure on the far right (ID 0-4: \textcolor{red}{red} team, ID 5-9: \textcolor{blue}{blue} team).}
  \label{fig:group_result}
\end{figure}

\begin{figure*}[t]
\centering
  \includegraphics[width=\columnwidth]{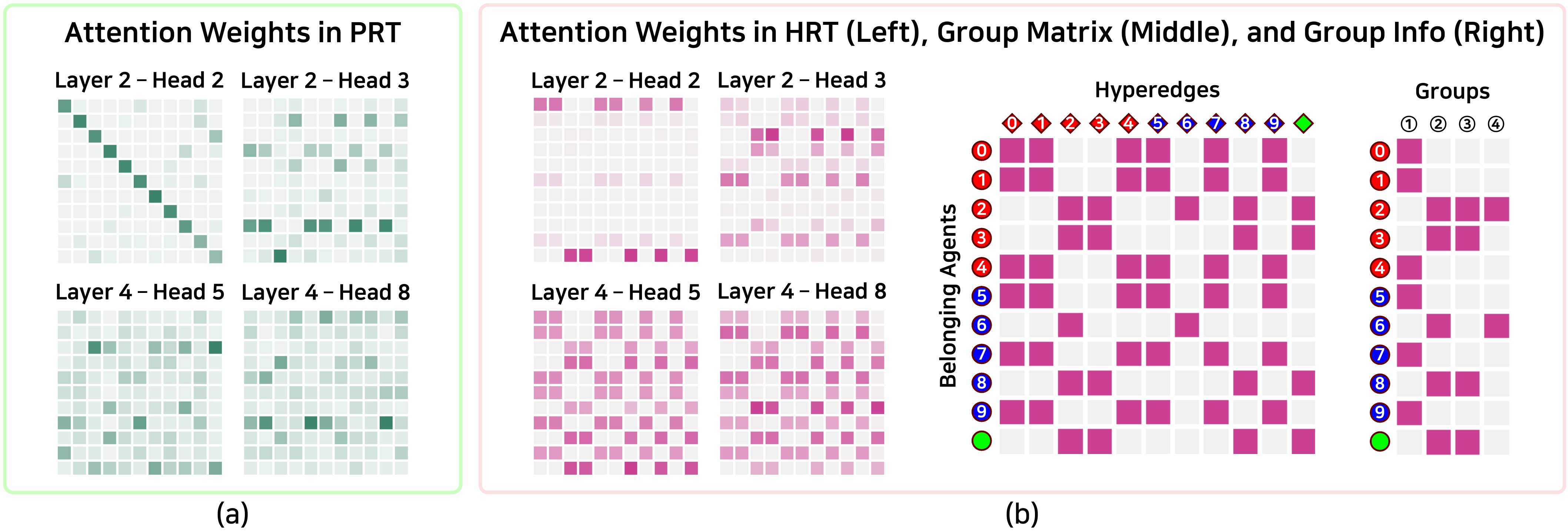}
  \caption{\textbf{(a) Visualization of attention weights in PRT. (b) Visualization of attention weights in HRT (left), the group incidence matrix $\mathbf{G}$ (middle), and the corresponding groups (right).} The sample results from the identical test data sample used in Figure \ref{fig:group_result}. For the attention visualization, min-max normalization is applied to ensure a clearer visual understanding, and darker colors indicate higher attention weights.}
  \label{fig:attention}
\end{figure*}

\noindent\textbf{Group Estimation.}\hspace{0.4em}
Figure \ref{fig:group_result} displays examples of group estimation results on the NBA dataset, and Figure \ref{fig:attention}-(c) shows the matrix for unique groups. Each group reflects a specific strategy. For example, in Group \#1, defenders employ man-to-man marking on offensive players during zone defense (See supplementary material for details).

\noindent\textbf{Attention Weights Visualization.}\hspace{0.4em}
As illustrated in Figure \ref{fig:attention}-(a) and -(b), the attention weights of PRT and HRT focus on the pair-wise and group-wise relations, respectively. Notably, for HRT, the attention weights in HRT show the same accentuation as the group incidence matrix displayed on the right side of Figure \ref{fig:attention}-(b). Consequently, the visualization demonstrates that the proposed MARTE promotes the attention weights to emphasize interactions between agents, particularly focusing on group-wise relations.


\section{Conclusion}

\noindent\textbf{Conclusion.}\hspace{0.4em}
This study proposes MART, a multiscale relational transformer network for multi-agent trajectory prediction. MART considers interactions at both individual and group scales via the hypergraph transformer mechanism. Within the MART framework, we introduce the MARTE and AGE modules. MARTE extends the capabilities of the relational transformer and promotes attention weights to emphasize interactions between agents, particularly focusing on group-wise relations. The AGE module outperforms pioneering group estimation modules in terms of performance and efficiency. Our comprehensive evaluation across three real-world datasets demonstrates that MART achieves state-of-the-art performance in multi-agent trajectory prediction, showcasing its potential to advance the field.

\noindent\textbf{Limitation and Future Work.}\hspace{0.4em}
Our model does not fully account for the temporal context of trajectories because it incorporates the past trajectory into the agent embedding. In future work, we plan to extend the model to incorporate spatio-temporal relationships. In addition, we aim to explore a trajectory prediction model with multi-modal inputs (e.g., trajectory inputs with scene context) because scene context, such as maps or images, is important in path prediction.

\section*{Acknowledgements}
This research was supported by the Future Challenge Program through the Agency for Defense Development funded by the Defense Acquisition Program Administration (No. UC200015RD).

\bibliographystyle{splncs04}
\bibliography{main}

\input{suppl}

\end{document}

%% file: suppl.tex
\clearpage
\def\thesection{\Alph{section}}
\setcounter{section}{0}

\hspace*{\fill} \textbf{\Large Supplementary Material} \hspace*{\fill}

\section{Detailed Model Architecture}

We provide detailed descriptions of the components in the model architecture, which were briefly explained earlier. For clarity, the bias vectors are omitted in this study. The components are detailed as follows: the node/pair-wise edge/hyperedge feature initializer $\mathcal{F}_{\mathrm{NI}}(\cdot)$ / $\mathcal{F}_{\mathrm{PI}}(\cdot)$ / $\mathcal{F}_{\mathrm{HI}}(\cdot)$, the transformer-like node / pair-wise / hyperedge update function $\mathcal{F}_n(\cdot)$ / $\mathcal{F}_e(\cdot)$ / $\mathcal{F}_h(\cdot)$, and the decoder $\mathcal{F}_{\mathrm{D},k}(\cdot)$.

\begin{itemize}
    \item $\mathcal{F}_{\mathrm{NI}}(\cdot)$: It comprises two fully connected layers. The sinusoidal positional encoding \cite{vaswani2017attention} blends the temporal information before the last layer. The equations are as follows:
    \begin{equation}
        \mathbf{n}^{(0)}_i = \mathrm{Flatten}(\mathrm{PosEnc}(\mathbf{X}_i\mathbf{W}_{\mathrm{NI,1}}))\mathbf{W}_{\mathrm{NI,2}},
    \end{equation}
    where $\mathrm{Flatten}(\cdot)$ denotes the flattening operation that transforms a multi-dimensional tensor into a one-dimensional tensor. $\mathrm{PosEnc}(\cdot)$ denotes the sinusoidal positional encoding \cite{vaswani2017attention} along the temporal dimension. $\mathbf{W}_{\mathrm{NI,1}}\in\mathbb{R}^{d_{in} \times d_n}$ and $\mathbf{W}_{\mathrm{NI,2}}\in\mathbb{R}^{T_p d_n \times d_n}$ are weight matrices for node feature initialization.
    
    \item $\mathcal{F}_{\mathrm{PI}}(\cdot)$ \& $\mathcal{F}_{\mathrm{HI}}(\cdot)$: Two layers of MLP with ReLU activation and the hidden dimension $d_h$.

    \item $\mathcal{F}_n(\cdot)$: Assume that the output of multi-head relational or hyper-relational attention for the $i$-th agent is $\mathbf{a}^{(l)}_{i}$, and given the node feature for the $i$-th agent $\mathbf{n}^{(l)}_{i}$, the updated node feature $\mathbf{n}^{(l+1)}_{i}$ via $\mathcal{F}_n(\cdot)$ is described as follows:
    \begin{equation}
    \begin{array}{l}
         \mathbf{u}^{(l)}_{i} = \mathrm{LayerNorm}(\mathbf{a}_{i}^{(l)}\mathbf{W}_1 + \mathbf{n}_{i}^{(l)}) \in \mathbb{R}^{d_n}, \\
         \mathbf{z}^{(l)}_{i} = \mathrm{FeedForward}(\mathbf{u}^{(l)}_{i}) \in \mathbb{R}^{d_n}, \\
         \mathbf{n}^{(l+1)}_{i} = \mathrm{LayerNorm}(\mathbf{z}_{i}^{(l)} + \mathbf{u}_{i}^{(l)}) \in \mathbb{R}^{d_n},
    \end{array}
    \end{equation}
    where $\mathbf{W}_1 \in \mathbb{R}^{d_n \times d_n}$ is a weight matrix, $\mathrm{LayerNorm}(\cdot)$ denotes the layer normalization \cite{ba2016layer}, and $\mathrm{FeedForward}(\cdot)$ represents two layers of MLP with ReLU activation and a hidden dimension of $d_h$.

    \item $\mathcal{F}_e(\cdot)$ \& $\mathcal{F}_h(\cdot)$: Given the message vector $\mathbf{m}^{(l)}_{i}$ for the $i$-th edge feature $\mathbf{e}^{(l)}_{i}$, transformer-like edge update function is described as follows:
    \begin{equation}
    \begin{array}{l}
         \mathbf{u}'^{(l)}_{i} = \mathrm{LayerNorm}(\mathbf{m}_{i}^{(l)}\mathbf{W}_2 + \mathbf{e}_{i}^{(l)}) \in \mathbb{R}^{d_e}, \\
         \mathbf{z}'^{(l)}_{i} = \mathrm{FeedForward}(\mathbf{u}'^{(l)}_{i}) \in \mathbb{R}^{d_e}, \\
         \mathbf{e}^{(l+1)}_{i} = \mathrm{LayerNorm}(\mathbf{z}'^{(l)}_{i} + \mathbf{u}'^{(l)}_{i}) \in \mathbb{R}^{d_e},
    \end{array}
    \end{equation}
    where $\mathbf{W}_2 \in \mathbb{R}^{d_h \times d_e}$ is a weight matrix.
    
    \item $\mathcal{F}_{\mathrm{D,k}}(\cdot)$: Three layers of MLP with ReLU activation and hidden dimensions of $[d_\mathrm{D}, d_\mathrm{D} / 2]$, where $d_\mathrm{D}$ denotes the decoder dimension.
\end{itemize}

\section{Further Experimental Details}\label{sec:Implementation_details}
\subsection{Datasets}
\textbf{NBA SportVU Dataset (NBA).} The NBA collected the NBA SportVU dataset using the SportVU tracking system during the 2015–2016 NBA season. The raw data is in JSON format and contains frame information, such as the IDs and positions of the players on the court, team IDs, and the game period. Consequently, the trajectories of the 10 players and the ball during actual basketball games were recorded and included in the dataset. Following \cite{xu2022dynamic, mao2023leapfrog}, we used 32.5k samples for training, 5.0k samples for validation, and 12.5k samples for testing. Additionally, we utilized the past 2.0 seconds (10 frames) of trajectories to forecast the next 4.0 seconds (20 frames).

\textbf{ETH-UCY dataset.} The ETH-UCY dataset \cite{pellegrini2009you, lerner2007crowds} includes five subsets: ETH, HOTEL, UNIV, ZARA1, and ZARA2, each comprising various pedestrian motion scenes captured at a frame rate of 2.5Hz. Based on previous studies \cite{alahi2016social, gupta2018social, bae2022gpgraph}, our model predicts the future trajectories of multiple agents for 4.8 s (12 frames) using the given 3.2 s (8 frames) of past trajectories. For evaluation, we train on four sets and assess the performance on the unused set, utilizing the leave-one-out method.

\textbf{Stanford Drone Dataset (SDD).} The SDD dataset \cite{robicquet2016learning} is a large-scale pedestrian motion dataset, recorded from a bird's eye view on a university campus and consisting of 20 scenes. We utilize the standard train-test split based on a previous study \cite{mangalam2020not}. Then, given the 3.2 seconds (8 frames) of past trajectories, we predict the future trajectories for 4.8 seconds (12 frames).

\subsection{Baseline Methods}
We describe the baselines compared in Section \ref{subsec:comparison_sota}.
\begin{itemize}
    \item \textbf{NMMP} \cite{hu2020collaborative}: This method introduces neural motion message-passing to model interactions among traffic actors for motion prediction in vehicles and social robots.
    \item \textbf{STAR} \cite{yu2020spatio}: This model is a spatio-temporal graph transformer focusing on trajectory prediction for crowd motion dynamics by leveraging attention mechanisms.
    \item \textbf{PECNet} \cite{mangalam2020not}: The model introduces the predicted endpoint-conditioned network for human trajectory prediction. This method models the stochastic endpoints and proposes a novel non-local social pooling layer.
    % \item \textbf{Trajectron++} \cite{salzmann2020trajectron++}: This method is a modular, graph-structured recurrent model to predict the trajectories, integrating dynamics and heterogeneous data.
    % \item \textbf{EvolveGraph} \cite{li2020evolvegraph}: This approach introduces a generic trajectory prediction model that recognizes explicit relational structures and predicts agent interactions.
    % \item \textbf{Agentformer} \cite{yuan2021agentformer}: A transformer-based multi-agent trajectory prediction model that considers time and social dimensions through agent-aware attention.
    \item \textbf{GroupNet} \cite{xu2022groupnet}: This method proposes a multiscale hypergraph message-passing neural network for multi-agent trajectory prediction to capture individual and group behaviors.
    \item \textbf{MemoNet} \cite{xu2022remember}: This model is a novel instance-based trajectory prediction framework designed to predict future intentions by establishing direct connections between present scenarios and previously observed instances.
    \item \textbf{MID} \cite{gu2022stochastic}: This approach, termed motion indeterminacy diffusion (MID), employs a parameterized Markov chain and a transformer-based diffusion model to progressively refine trajectory predictions.
    \item \textbf{NPSN} \cite{bae2022non}: This method addresses stochastic pedestrian trajectory prediction variability by introducing the Quasi-Monte Carlo method for improved sampling and proposing the Non-Probability Sampling Network (NPSN).
    \item \textbf{GP-Graph} \cite{bae2022gpgraph}: GP-Graph is a pedestrian trajectory prediction model that captures intra-group and inter-group interactions by utilizing a group assignment module to cluster pedestrians into proximity-based groups.
    \item \textbf{DynGroupNet} \cite{xu2022dynamic}: This method, an extension of GroupNet \cite{xu2022groupnet}, is a dynamic-group-aware network designed to enhance the modeling of interactions among agents for trajectory prediction, capable of capturing time-varying interactions at both the pair-wise and group scales.
    \item \textbf{ET} \cite{bae2023eigentrajectory}: The paper presents EigenTrajectory (ET), a method for enhancing pedestrian trajectory prediction by creating a compact ET space for pedestrian movement representations.
    \item \textbf{EqMotion} \cite{xu2023eqmotion}: EqMotion is a motion prediction model that is equivariant under Euclidean geometric transformations while considering invariant interaction reasoning.
    \item \textbf{LED} \cite{mao2023leapfrog}: LED is a diffusion-based trajectory prediction model that offers real-time, accurate, and diverse predictions. This method accelerates inference speed by bypassing numerous denoising steps with a leapfrog initializer.
\end{itemize}

\subsection{Model Input}
We utilize the absolute and relative positions as model inputs for the NBA dataset, resulting in an input dimension $d_{in}$ of 4. For the ETH-UCY and SDD datasets, the relative positions are used as input, establishing the input dimension $d_{in}$ at 2.

\subsection{Model Hyperparameter Search}

\begin{table}[t]
\setlength\tabcolsep{23.3pt} % let LaTeX compute intercolumn whitespace
\centering
\caption{Hyperparameter candidates within the model architecture. \textbf{Bold} denotes the selected hyperparameters.} \label{tab:suppl_hyperparameter}
\begin{tabular*}{\columnwidth}{c|c|c}
\toprule
\textbf{Parameter Name} & \textbf{Notation} & \textbf{Considered Values} \\
\midrule
node dimension & $d_n$ & $\{\textbf{64}, 128, 192\}$ \\
edge dimension & $d_e$ & $\{\textbf{64}, 128, 192\}$ \\
hidden dimension & $d_h$ & $\{64, \textbf{128}, 192\}$ \\
decoder dimension & $d_{\mathrm{D}}$ & $\{\textbf{128}, 256, 384, 512\}$ \\
number of heads & - & $\{1, 2, 4, \textbf{8}\}$ \\
\bottomrule
\end{tabular*}
\end{table}

Table \ref{tab:suppl_hyperparameter} lists the hyperparameters considered within the model architecture. Utilizing these candidates, we conducted a hyperparameter search on the NBA dataset, with the number of layers ($L$) fixed at 1. Subsequently, the optimal value for $L$ was established, as detailed in Section \ref{subsec:suppl_num_layers}.

\subsection{Model Training} \label{subsec:suppl_model_training}

\begin{table}[t]
\setlength\tabcolsep{14.65pt} % let LaTeX compute intercolumn whitespace
\footnotesize\centering
\caption{Learning rate, batch size, and number of training epochs used in this study.} \label{tab:suppl_model_training}
\resizebox{\columnwidth}{!}{
\begin{tabular*}{1.2\columnwidth}{c|c|c|c|c}
\toprule
\textbf{Dataset} & \textbf{Subset} & \textbf{Learning Rate} & \textbf{Batch Size} & \textbf{Training Epochs} \\
\midrule
NBA & - & $5.0\times10^{-4}$ & 64 & 100 \\
\midrule
SDD & - & $1.0\times10^{-3}$ & 256 & 300 \\
\midrule
\multirow{5}{*}{ETH-UCY} & ETH & $1.0\times10^{-3}$ & \multirow{5}{*}{64} & \multirow{5}{*}{300} \\
 & HOTEL & $1.8\times10^{-3}$ & & \\
 & UNIV & $1.0\times10^{-3}$ & & \\
 & ZARA1 & $1.2\times10^{-3}$ & & \\
 & ZARA2 & $1.2\times10^{-3}$ & & \\
\bottomrule
\end{tabular*}
}
\end{table}

We used an Adam optimizer \cite{kingma2014adam} to train our network. The specific training hyperparameters, including the learning rate, batch size, and the number of training epochs, are detailed in Table \ref{tab:suppl_model_training}. We applied a learning rate decay of 0.5 per 100 epochs for all datasets. Our training and evaluation were conducted on an NVIDIA GeForce RTX 3090. Our implementations were on PyTorch 2.1.0 \cite{paszke2019pytorch}.

\section{Supplementary Experiments}

\subsection{Effect of the Number of Layers}\label{subsec:suppl_num_layers}
Table \ref{tab:suppl_num_layers} displays the effect of the number of layers ($L$) in MART. This experiment was conducted on the NBA dataset using the same condition described in Section \ref{subsec:suppl_model_training}. The performance of MART gradually improves until $L=4$, after which it begins to decline. This experiment illustrates why we selected four layers.

\begin{table}[t]
    \begin{minipage}{.49\linewidth}
      \caption{Effect of the number of layers ($L$) in MARTE on the NBA datsaet. The reported values are in meters.} \label{tab:suppl_num_layers}
      \setlength\tabcolsep{18.1pt}
      \centering
        \resizebox{\columnwidth}{!}{
\begin{tabular*}{1.2\columnwidth}{c|cc}
\toprule
$L$ & $\min\mathrm{ADE}_{20}$ & $\min\mathrm{FDE}_{20}$ \\
\midrule
1 & 0.761 & 0.943 \\
2 & 0.738 & 0.916 \\
3 & 0.733 & 0.908 \\
\textbf{4} & \textbf{0.727} & \textbf{0.903} \\
5 & 0.741 & 0.919 \\
\bottomrule
\end{tabular*}
        
        }

    \end{minipage}%
    \hspace{0.4em}
    \begin{minipage}{.49\linewidth}
    \caption{Ablation on the NBA dataset for the estimated gradient within the AGE module. Function types (a), (b), and (c) match with Figure \ref{fig:suppl_ste}(a), (b), and (c), respectively.} \label{tab:suppl_ste}
    \setlength\tabcolsep{10.3pt}
    \centering
        \resizebox{\columnwidth}{!}{
\begin{tabular*}{1.2\columnwidth}{c|cc}
\toprule
\textbf{Func. Type} & $\min\mathrm{ADE}_{20}$ & $\min\mathrm{FDE}_{20}$ \\
\midrule
(a) & 0.733 & 0.906 \\
(b) & \textbf{0.727} & \textbf{0.903} \\
(c) & 0.731 & 0.907 \\
\bottomrule
\end{tabular*}
}
\end{minipage} 
\end{table}

\begin{figure}[h]
\centering
  \includegraphics[width=0.7\columnwidth]{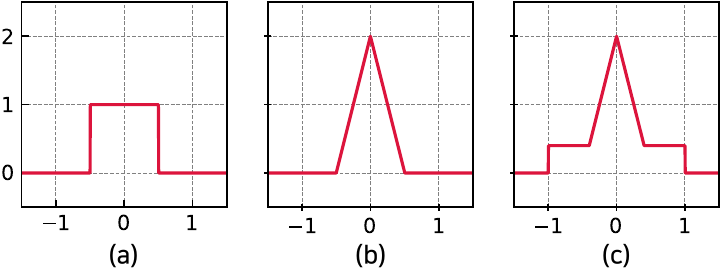}
  \caption{\textbf{Candidates for the estimated gradient in AGE.} (a) Modified STE \cite{hubara2016binarized}. (b) Modified triangle-shaped estimator \cite{liu2018bi}. (c) Long-tailed estimator \cite{xu2019accurate}.}
  \label{fig:suppl_ste}
\end{figure}

% \begin{table}[h]
% \setlength\tabcolsep{6.2pt} % let LaTeX compute intercolumn whitespace
% \footnotesize\centering
% \caption{Effect of the number of layers ($L$) in MARTE. This experiment was conducted on the NBA dataset. The reported values are $\min\mathrm{ADE}_{20}$/$\min\mathrm{FDE}_{20}$ in meters.} \label{tab:suppl_num_layers}
% \begin{tabular*}{\columnwidth}{c|cccc}
% \toprule
% \multirow{2}{*}{$L$\vspace{-0.6em}} & \multicolumn{4}{c}{\textbf{Prediction Time}} \\
% \cmidrule{2-5}
% & 1.0s & 2.0s & 3.0s & 4.0s \\
% \midrule
% 1 & 0.184 / 0.272 & 0.369 / 0.534 & 0.562 / 0.750 & 0.761 / 0.943 \\
% 2 & 0.178 / 0.258 & 0.356 / 0.512 & 0.543 / 0.721 & 0.738 / 0.916 \\
% 3 & 0.181 / 0.260 & 0.356 / 0.507 & 0.541 / 0.717 & 0.733 / 0.908 \\
% \textbf{4} & \textbf{0.177} / \textbf{0.255} & \textbf{0.352} / \textbf{0.501} & \textbf{0.535} / \textbf{0.708} & \textbf{0.727} / \textbf{0.903} \\
% 5 & 0.180 / 0.262 & 0.359 / 0.513 & 0.547 / 0.723 & 0.741 / 0.919 \\
% \bottomrule
% \end{tabular*}
% \end{table}

\subsection{Ablation for Estimated Gradient within AGE} \label{subsec:suppl_function}
Figure \ref{fig:suppl_ste} illustrates the candidates for the estimated gradient within AGE. Table \ref{tab:suppl_ste} lists the ablations for the estimated gradient within AGE. The modified triangular-shaped estimator performs better than others. The original STE \cite{hubara2016binarized} and the triangle-shaped estimator \cite{liu2018bi} approximate the gradient of $\mathrm{Sign}(\cdot)$. Because we estimate the unit step function $\mathcal{U}(\cdot)$, we have made slight modifications to them, as illustrated in Figures \ref{fig:suppl_ste}-(a) and -(b).

\section{Further Qualitative Results}

\subsection{Trajectory Prediction}
\begin{figure*}[!t]
  \includegraphics[width=\columnwidth]{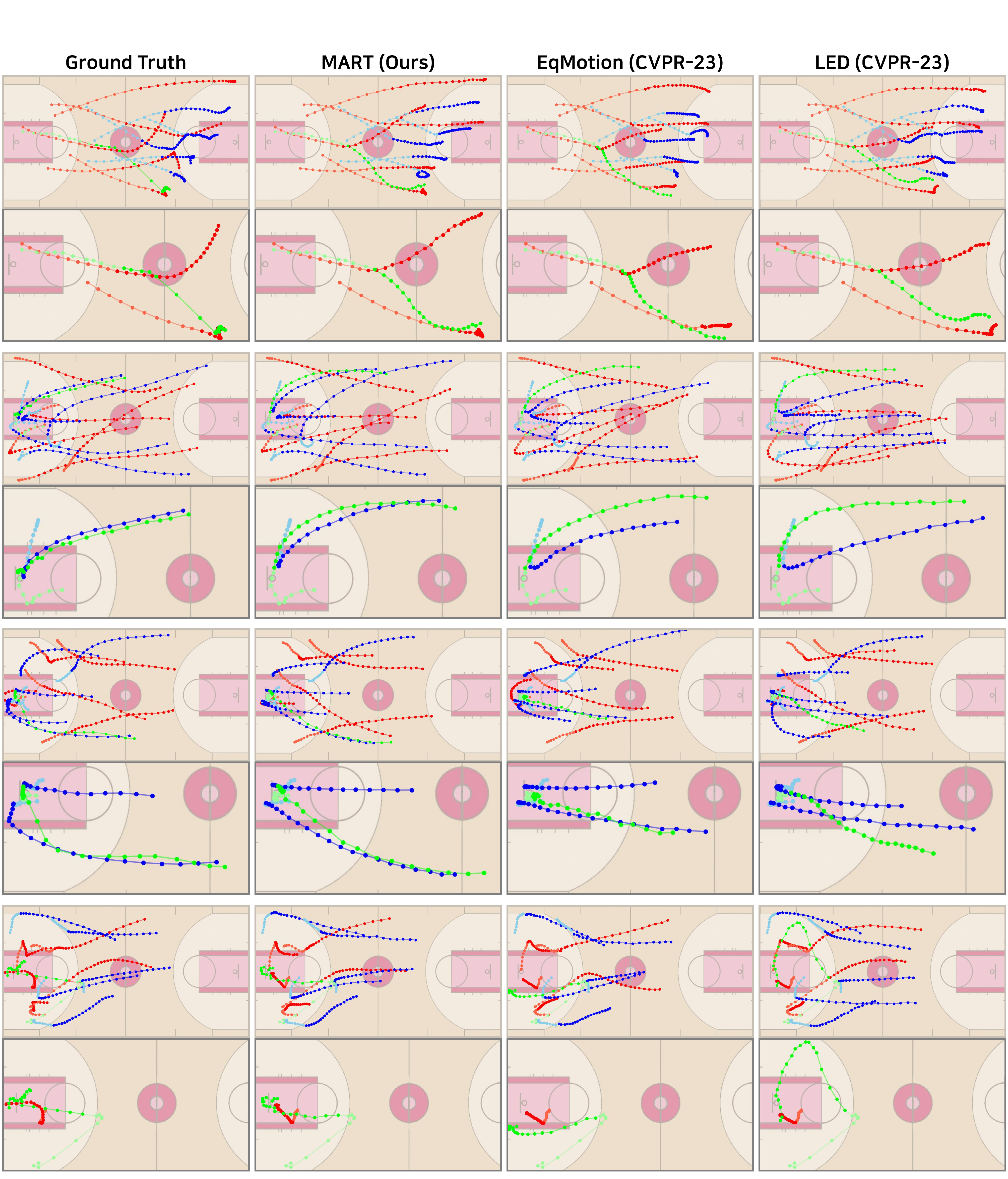}
  \caption{\textbf{Further qualitative result comparisons on NBA dataset.} We qualitatively compare our method with two recent state-of-the-art methods using the best of 20 predictions. The khaki border illustrates predictions for all agents, while the gray border focuses on the ball-keeping player and the basketball. When comparing our method across more samples, it predicts more accurate future trajectories compared to LED \cite{mao2023leapfrog} and EqMotion \cite{xu2023eqmotion}. Notably, as illustrated in figures with a khaki border, MART provides robust predictions about the ball-keeping player and the basketball, whereas other methods do not. These results demonstrate that MART can handle more complex interactions compared to state-of-the-art (SOTA) methods. (Light color represents the past trajectory, while \textcolor[HTML]{0000FF}{blue}/\textcolor[HTML]{FF0000}{red}/\textcolor[HTML]{00FF00}{green} indicate the two teams and basketball.)}
  \label{fig:suppl_nba}
\end{figure*}

\begin{figure*}[!t]
  \includegraphics[width=\columnwidth]{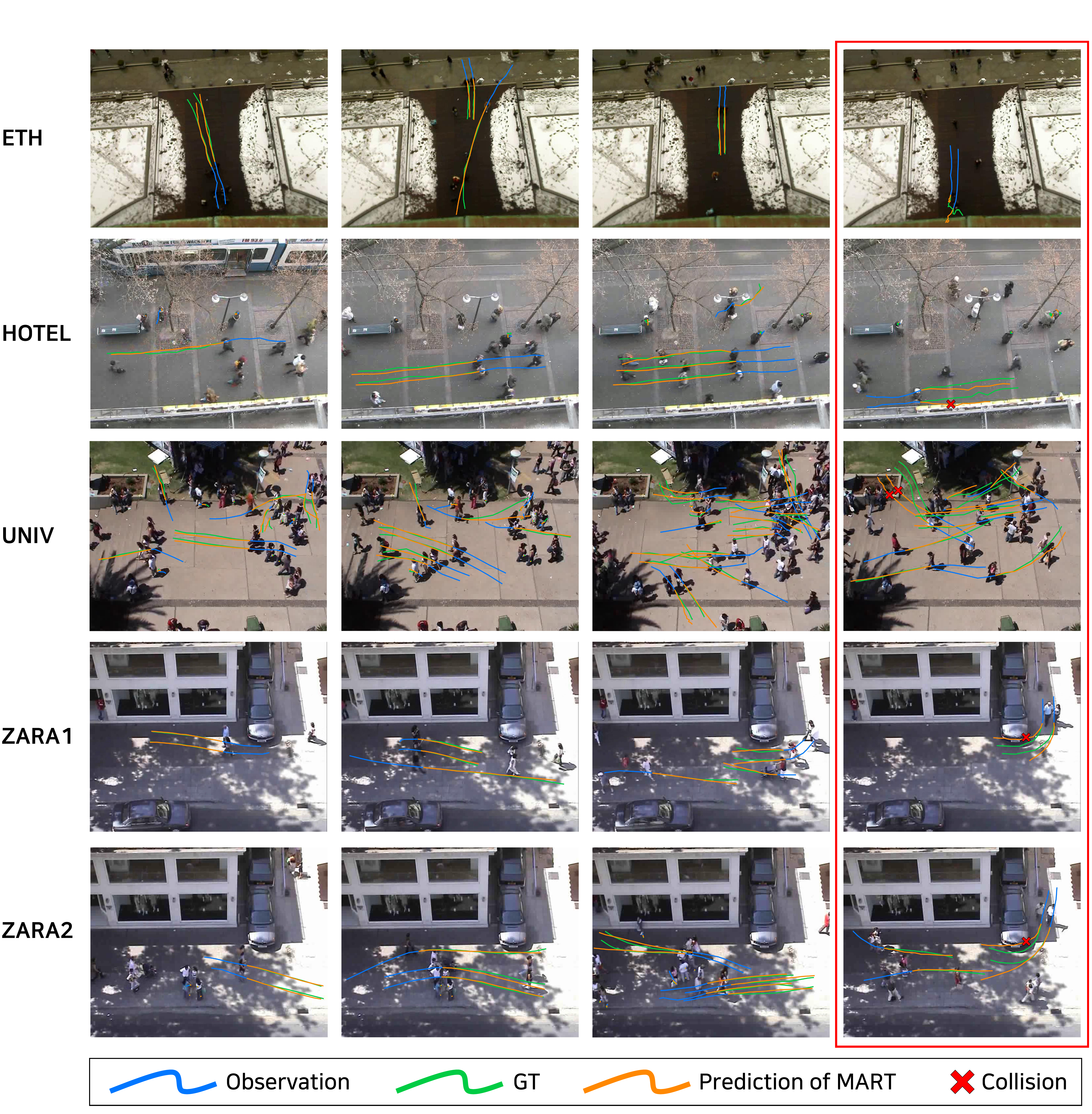}
  \caption{\textbf{Examples of trajectory prediction results on the ETH-UCY dataset.} The figure displays trajectory prediction results of MART with the best of 20 predictions. The column boxed in red (\textcolor{red}{\Squarepipe}) indicates bad examples, where the predicted path collides with obstacles such as walls, flower beds, and cars.}
  \label{fig:suppl_eth}
\end{figure*}

Figure \ref{fig:suppl_nba} presents additional qualitative result comparisons on the NBA dataset. As depicted in the figure, our method yields more accurate predictions than recent state-of-the-art (SOTA) methods, including LED \cite{mao2023leapfrog} and EqMotion \cite{xu2023eqmotion}, across a larger number of samples. Figure \ref{fig:suppl_eth} shows qualitative results of MART on the ETH-UCY dataset. In some cases, we observed that the predicted trajectory does not satisfy the scene context, such as obstacles and maps.

\subsection{Group Estimation}

\begin{figure*}[t]
  \includegraphics[width=\columnwidth]{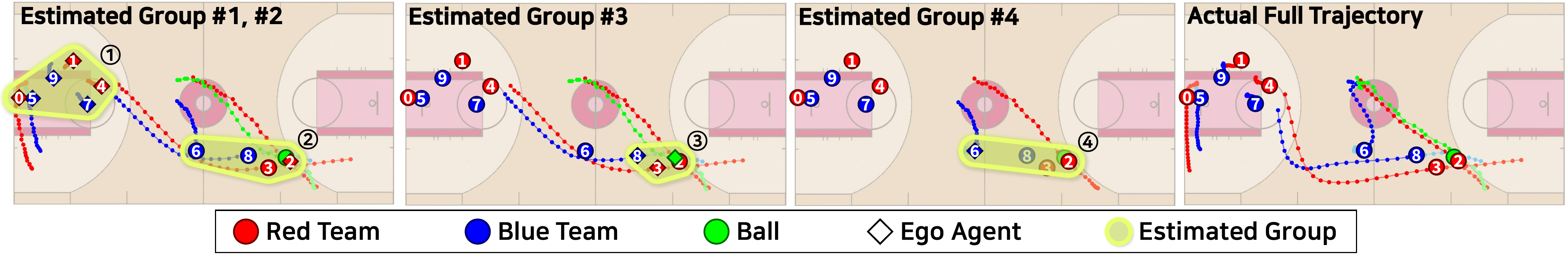}
  \caption{\textbf{Visualization of group estimation result.} The left side of the court represents the \textcolor{blue}{blue} team's region, while the right side of the court represents the \textcolor{red}{red} team's region. Group IDs are denoted as \textcircled{\raisebox{-0.9pt}{1}}, \textcircled{\raisebox{-0.9pt}{2}}, \textcircled{\raisebox{-0.9pt}{3}}, and \textcircled{\raisebox{-0.9pt}{4}}. Player IDs are provided in the figure on the far right (ID 0-4: \textcolor{red}{red} team, ID 5-9: \textcolor{blue}{blue} team).}
  \label{fig:suppl_group}
\end{figure*}

\begin{figure}[t]
\begin{center}
  \includegraphics[width=0.6\columnwidth]{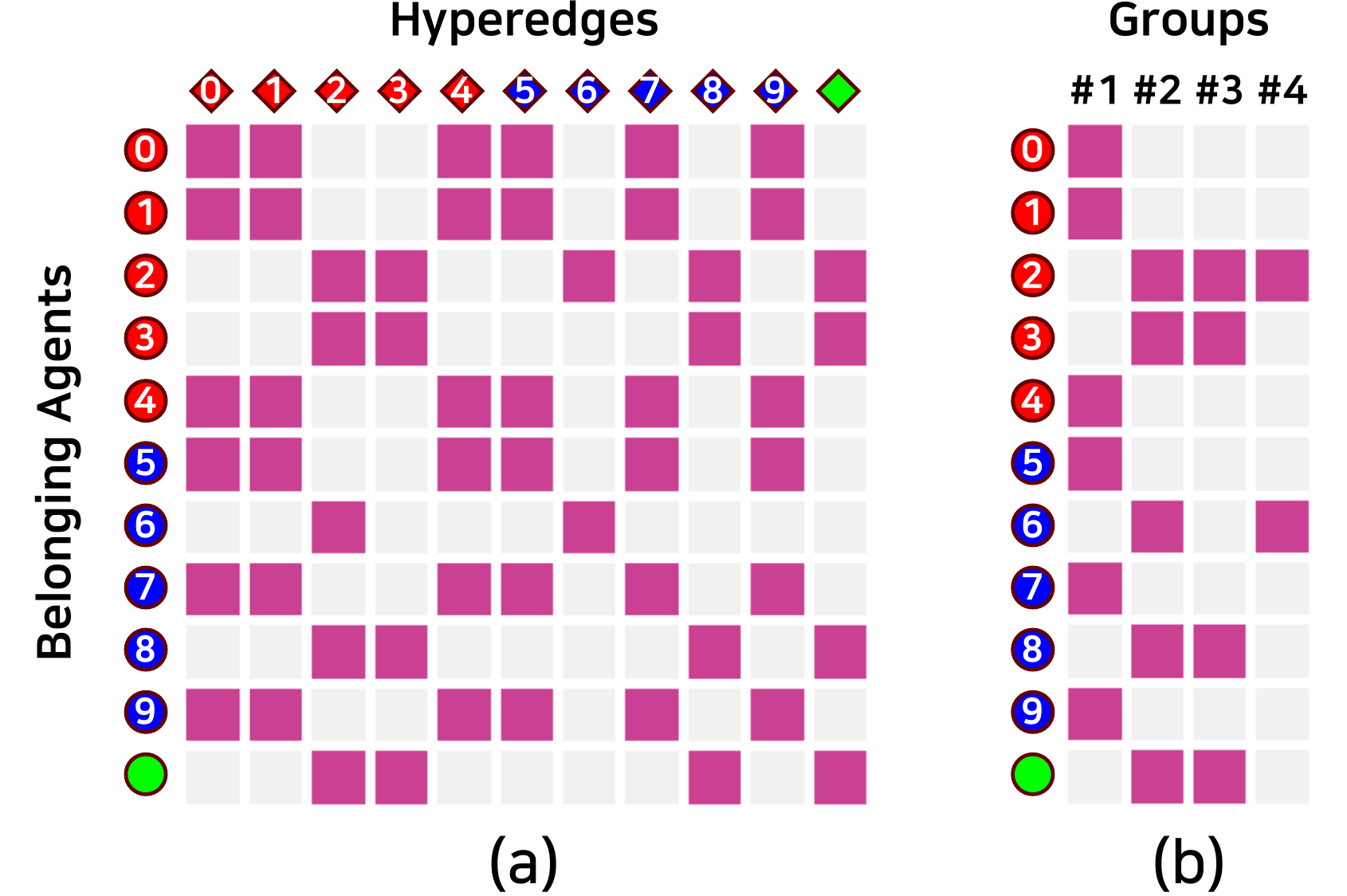}
  \caption{\textbf{Group estimation result.} (a) Visualization of the group incidence matrix $\mathbf{G}$ (b) Visualization of unique groups (i.e., the group incidence matrix that replicated groups excluded). Both resulted from the identical test data sample used in Figure \ref{fig:suppl_group}.}\label{fig:suppl_hyperedge}
\end{center}
\end{figure}

This section provides a detailed description of the group estimation results presented in Section \ref{subsec:qualitative_results}. Figure \ref{fig:suppl_group} displays the visualization of group estimation results on the NBA dataset, while Figure \ref{fig:suppl_hyperedge} shows the visualization of the group incidence matrix $\mathbf{G}$ and the unique groups corresponding to Figure \ref{fig:suppl_group}. Each group reflects a specific strategy as follows:
\begin{itemize}
    \item \textbf{Group \#1 (Zone Defense Group)}: Blue team defenders with player IDs 5, 7, and 9 employ man-to-man marking on red team offensive players with player IDs 0, 4, and 1, respectively.
    \item \textbf{Group \#2 (Global Att\&Def Group)}: The red team player with ID 2 dribbles the ball while being marked by the blue team player with ID 6. The player with ID 3 from the red team is connected to the player with ID 2 through an attacking link. Blue team players with IDs 6 and 8 are marking the red team players with IDs 2 and 3, respectively.
    \item \textbf{Group \#3 (Local Att\&Def Group)}: The red team player with ID 2 is dribbling with the ball. The blue team player with ID 8 is marking the attacking red team player with ID 3.
    \item \textbf{Group \#4 (Local Att\&Def Group)}: The blue team player with ID 6 defends against the attacking red team player with ID 2.
\end{itemize}

